\documentclass[twocolumn,journal]{IEEEtran}
\usepackage[T1]{fontenc}
\usepackage{units}
\usepackage{amsmath}
\usepackage{amssymb}
\usepackage{graphicx}
\usepackage[unicode=true,
 bookmarks=true,bookmarksnumbered=true,bookmarksopen=true,bookmarksopenlevel=1,
 breaklinks=false,pdfborder={0 0 0},backref=false,colorlinks=false]
 {hyperref}
\hypersetup{pdftitle={Your Title},
 pdfauthor={Your Name},
 pdfpagelayout=OneColumn, pdfnewwindow=true, pdfstartview=XYZ, plainpages=false}

\makeatletter

\providecommand{\tabularnewline}{\\}

 \let\oldforeign@language\foreign@language
 \DeclareRobustCommand{\foreign@language}[1]{%
   \lowercase{\oldforeign@language{#1}}}

\usepackage[caption=false,font=footnotesize]{subfig}

\makeatother

\begin{document}

\title{Identification of refugee influx patterns in Greece via model-theoretic
analysis of daily arrivals}

\author{Harris~V.~Georgiou~(MSc, PhD)\thanks{Harris~Georgiou is an associate post-doc researcher with the Signal
\& Image Processing Lab (SIPL), Department of Informatics \& Telecommunications
(DIT), National Kapodistrian University of Athens (NKUA/UoA), Greece.\protect \\
E-mail: \protect\href{mailto:harris@xgeorgio.info}{harris@xgeorgio.info}
-- URL: \protect\href{http://xgeorgio.info}{http://xgeorgio.info}}}

\IEEEspecialpapernotice{Part I: Time frame Oct.2015\ --\ Jan.2016\\
Last updated: 7-May-2016}

\markboth{Ref.No: HG/GT.0507.01v1 -- Licensed under Creative Commons (BY-NC-SA)
4.0~\copyright~2016 Harris Georgiou}{}
\maketitle
\begin{abstract}
The refugee crisis is perhaps the single most challenging problem
for Europe today. Hundreds of thousands of people have already traveled
across dangerous sea passages from Turkish shores to Greek islands,
resulting in thousands of dead and missing, despite the best rescue
efforts from both sides. One of the main reasons is the total lack
of any early warning/alerting system, which could provide some preparation
time for the prompt and effective deployment of resources at the ``hot''
zones. This work is such an attempt, the first completely data-driven
study for a systemic analysis of the refugee influx in Greece, aiming
at: (a) the statistical and signal-level characterization of the smuggling
networks and (b) the formulation and preliminary assessment of such
models for predictive purposes, i.e., as the basis of such an early
warning/alerting protocol. To our knowledge, this is the first-ever
attempt to design such a system, since this refugee crisis itself
and its geographical properties are unique (intense event handling,
little/no warning). The analysis employs a wide range of statistical,
signal-based and matrix factorization (decomposition) techniques,
including linear \& linear-cosine regression, spectral analysis, ARMA,
SVD, Probabilistic PCA, ICA, K-SVD for Dictionary Learning, as well
as fractal dimension analysis. It is established that the behavioral
patterns of the smuggling networks closely match (as expected) the
regular ``burst'' and ``pause'' periods of store-and-forward networks
in digital communications. There are also major periodic trends in
the range of 6.2-6.5 days and strong correlations in lags of four
or more days, with distinct preference in the Sunday/Monday 48-hour
time frame. These results show that such models can be used successfully
for short-term forecasting of the influx intensity, producing an invaluable
operational asset for planners, decision-makers and first-responders.\end{abstract}


\IEEEpeerreviewmaketitle{}

\section{Introduction}

\IEEEPARstart{S}{ince} early January 2015, Europe has witnessed
an unprecedented influx of refugees from regions of war and conflict
in the Middle East, primarily Syria, Afghanistan and Iraq. For more
than 17 months now, Greece has become the main entrance gateway for
hundreds of thousands of people trying to get to central and northern
European countries. According to the United Nations High Commissioner
for Refugees (UNHCR) \cite{UNHCR-data3-url}, the International Organization
for Migration (IOM) \cite{IOM-data-url} and the Medecins Sans Frontieres
(MSF) \cite{MSF-rep-url}, during the year 2015 alone, more than a
million people reached Europe from Turkey and North Africa, seeking
safety and asylum.

In this context, the rapid allocation of proper resources is the most
critical factor in the success or failure of any rescue and relief
operations, especially in the ``hot'' zones. On the other hand,
the complete coverage of these areas of interest by patrols alone
is practically infeasible due the geographical extent of possible
landing points, as well as the excessive financial cost if these operations
are to be maintained on 24-hour rotation for many weeks and months.
Therefore, an early warning/alerting system for (expected) high refugee
influx, analogous to the ones used for extreme weather conditions
or possible wildfires in forests, would provide invaluable time for
the preparation and deployment of teams and equipment from staging
posts to specific areas of interest, promptly and effectively, in
order to save lives.

This study is the first (to our best knowledge) attempt for a completely
data-driven systemic analysis of the refugee influx data series, aiming
at: (a) the statistical and signal-level characterization of the smuggling
networks as a generating process; and (b) the draft formulation and
preliminary assessment of such models for predictive purposes, i.e.,
to produce short-term forecasting of the refugee influx, as part of
an early warning/alerting protocol. After the description of the material
(data series) used, a wide range of statistical, signal-based, spectral
and component analysis (decomposition) techniques are presented in
brief, each accompanied with the corresponding results and the conclusions
drawn from them. Finally, the general findings are further discussed
and future enhancements are proposed.

\subsection{Background}

There are only few passages across the sea borders between Greece
and Turkey where the distance is 5-7.5 n.m., hence these are the points
of interest for both the smuggling networks and the coast guard patrols.
At least 856,723 people came to Greece via Turkey, 80\% of which landed
at the island of Lesvos in the northern Aegean Sea. Nevertheless,
the lack of proper infrastructure, first-response coordination, early
warning and on-the-spot logistical support resulted in thousands of
casualties. The seer volume of the influx resulted in a total of 3,771
registered dead or missing persons in the Mediterranean Sea during
2015 \cite{UNHCR-portal-url,UNHCR-data3-url} , more than 832 in the
Aegean Sea. There were specific 24-hour time frames at the end of
October and the beginning of November 2015 when small beaches in the
northern shores of Lesvos, like the small port of Skala Sykamneas
with a population of only a few hundreds, received over 120 boats
landing there, each carrying 40-50 persons. 

During 2015, the Hellenic Coast Guard (HCG) has conducted over 6,300
Sea Search \& Rescue (SSAR) operations in these areas and more than
117,743 people have been rescued in the Aegean Sea \cite{HCG-data-url};
additionally, Turkish Coast Guard (TCG) has rescued at least 59,377
people, plus 339 dead or missing, in other incidents \cite{TCGC-data-url}.
Therefore, it is estimated that roughly 177,120 in 856,723 or more
than one in five people (1:4.84) coming across these passages ended
up rescued from the water; adding up the (estimated) dead and missing
\cite{IOM-data-url}, at least 832 in 177,952 or 1:214 (about two
persons every nine boats that sank) ended up dead or missing somewhere
in the Aegean Sea.

The first quarter of 2016 up the the first weeks of April resulted
in just over 178,882 arrivals in the Mediterranean Sea and another
737 dead or missing \cite{IOM-data-url}. More specifically, there
are 356 deaths in 24,581 arrivals in Italy and 375 deaths in 153,625
arrivals in Greece, the two major entrance points to Europe via the
Central/Eastern Med. routes (update: April 19th, 2016 \cite{IOM-flowrep-url}).
This yields a death ratio of 1:69 for Italy but 1:410 for Greece,
i.e., almost six times deadlier in comparison. 

It should also be noted that the deaths from sinking of migrant/refugee
ships inbound to Italy are usually underestimated due to the difficulties
in locating all the bodies in the open sea, as well as the under-reporting
of passengers on board. On April 18th 2016, a 30m boat capsized during
the night between Libya and Sicily, carrying an estimated 500-550
people; only 41 were rescued and a few dozen bodies were recovered
by the Italian Coast Guard (ICG) \cite{News-Apr16-url}. Similar major
events have occurred in the past, on October 3rd \& 11th 2013 (394+
dead/missing) and on April 20th 2014 (800 dead/missing), with survival
rates of less than 28\% \cite{News-Oct13-url,News-Apr15-url}. 

Operation \emph{Mare Nostrum} was a year-long naval and air mission
involving the Italian Coast Guard and Navy, after the major shipwrecks
of October 2013 in the Central Med. route from North Africa \cite{Oper-MareNostrum-url}.
During this period, more than 150,000 migrants and refugees were successfully
rescued in the sea. The operation ended in October 31st 2014 and it
was superseded by EU Frontex's \emph{Operation Triton} \cite{Oper-Triton-url},
another border patrol \& SSAR mission, with a much smaller force (on
a volunteer basis by 15 other European countries) and a much more
limited range in the same area. As a result, the incidents with hundreds
of people missing in the sea and the survival rates returned to their
prior states.

\begin{figure}[htbp]
\begin{centering}
\textsf{\includegraphics[width=8.5cm]{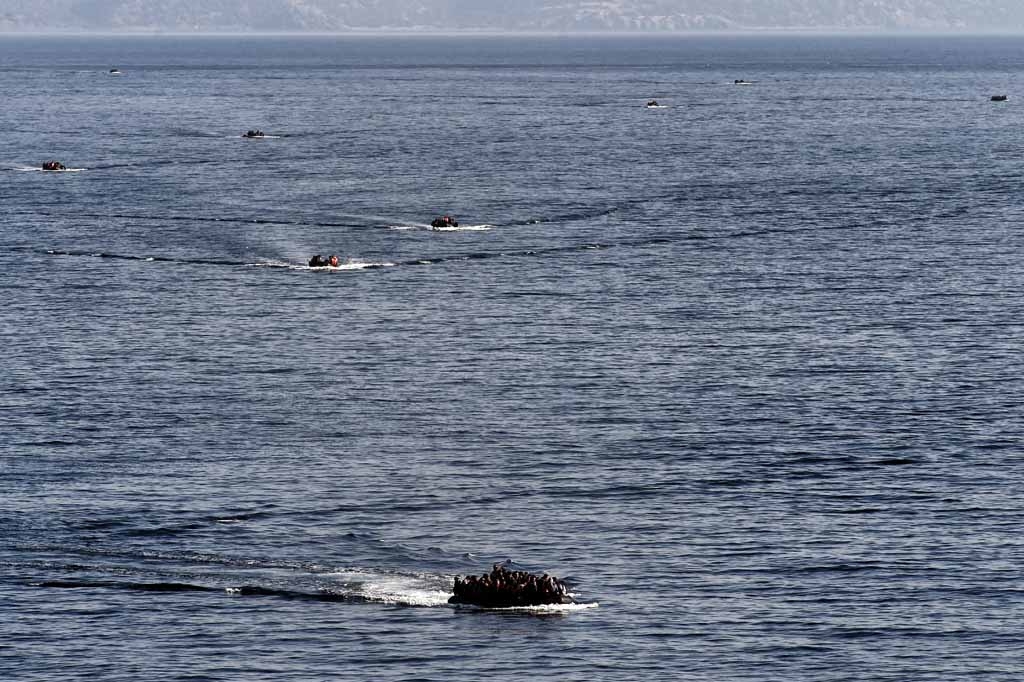}}
\par\end{centering}

\caption{\label{fig:Refugee-influx-snapshot-real}A snapshot photograph from
the northern beaches of Lesvos on October 2015, showing nine inflatable
boats carrying 40-50 each, heading to the landing zone with only minutes
apart (Credit: AFP / Aris Messinis).}
\end{figure}

\begin{figure}[htbp]
\begin{centering}
\textsf{\includegraphics[width=8.5cm]{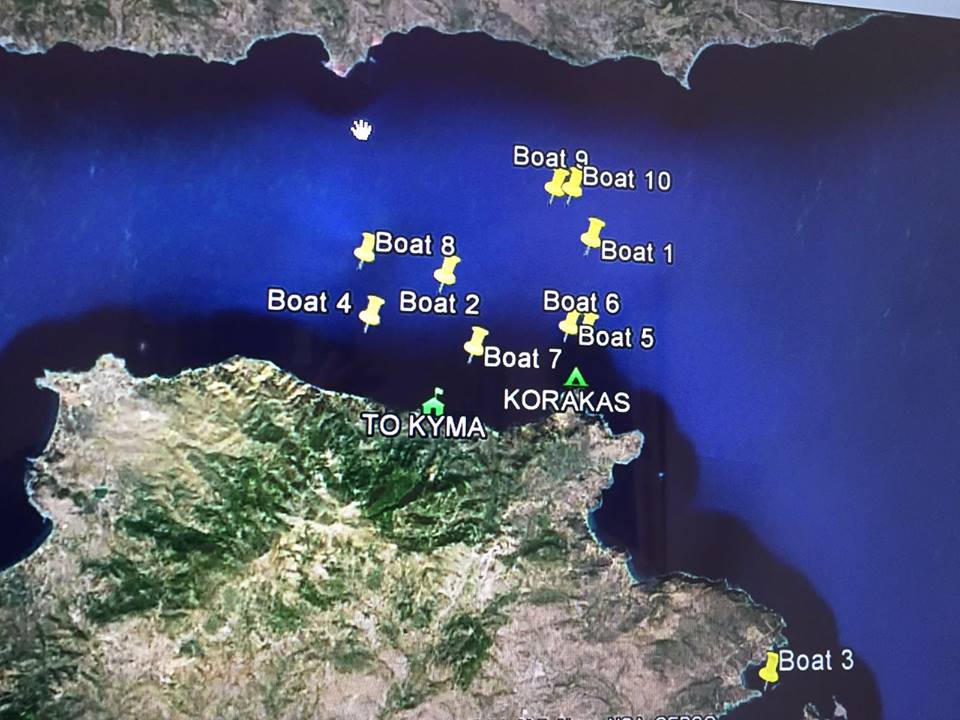}}
\par\end{centering}

\caption{\label{fig:Refugee-influx-snapshot-map}A snapshot from a live Google
map used by the SSAR elements in northern Lesvos, showing the identified
refugee boats heading towards the island on February 17th, 2016 (13:37'
local) (Credit: Proactiva Open Arms).}
\end{figure}

Despite the dreadful probabilities, migrants and refugees still prefer
to make their attempt via sea to Greece or Italy rather than via land
(mostly Turkish-Bulgarian borders) by a rate of more than 42:1, because
sea borders are inherently harder to patrol, fence or deny rescue.
This is a very clear explanation of why these narrow sea crosses between
Turkish sail-off beaches and the Greek islands received such a huge
volume of refugee influx, sometimes 5,000 people or more on a daily
basis, e.g. 12,000+ people within the 48 hours of 27-28 October 2015.
Figure \ref{fig:Refugee-influx-snapshot-real} and Figure \ref{fig:Refugee-influx-snapshot-map}
illustrate how this looks like in real life, with snapshots taken
on October 2015 (photo) and February 2016 (map). 

After the recent EU-Turkey deal in March 2016 \cite{News-EUdeal-url},
regarding the handling of asylum seekers and their mutual transfer
after proper registration, the influx to the EU is starting to shift
again from the Eastern (Greece) to the Central (Italy) Med. sea route,
resulting in a sharp rise of dead and missing ratios in boat sinking
incidents.

According to more recent numbers from IOM and MSF, by early May about
1,200 people have died in the Mediterranean Sea trying to reach Europe.
MSF, who has recently resumed its own SSAR operations in the area
between Libya and Italy, estimate that 976 people have died trying
to cross from Libya to Europe so far this year (early May), yielding
a death ratio of roughly 1:30. The boat trip from Libya to Italy is
much longer and perilous than the crossings from Turkey to Greece,
8-12 hours for about 150 n.m. rather than 25-40 minutes for 5-7.5
n.m. in comparison. As a result, massive boat sinking and capsizing
events every week are drastically increasing the death/missing total
and the true death ratio in the Central Med. route to Europe.

\subsection{Problem statement}

\begin{figure}[htbp]
\begin{centering}
\textsf{\includegraphics[width=8.5cm]{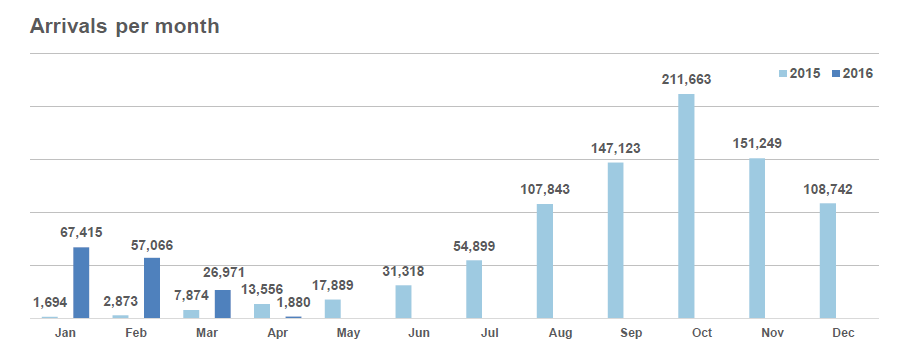}}
\par\end{centering}

\caption{\label{fig:Refugee-influx-statistics}Refugee influx statistics per
month, 2015 and early 2016 (Greece).}
\end{figure}

The most challenging task in managing such intense influx of refugees
in the very large coastline of Greece, in such a short notice due
to the very short distance from Turkey, is being the proper allocation
and rapid response of SSAR resources, as well as medical care at the
beaches and ports in case of shipwrecks. Unfortunately, neither the
resources nor the coordination was in place when it was needed the
most. Figure \ref{fig:Refugee-influx-statistics} shows the total
refugee influx intensity in all the Greek islands during 2015 and
the first months of 2016 \cite{UNHCR-portal-url}, peaking around
the period September-November 2015. During that time frame, at least
half of the rescues were conducted by volunteers, fishermen and other
non-tasked vessels, while first-response medical care was often performed
on the ground with little to no resources available to extremely limited
staff (primarily volunteer doctors and lifeguards).

\begin{figure}[htbp]
\begin{centering}
\textsf{\includegraphics[width=8.5cm]{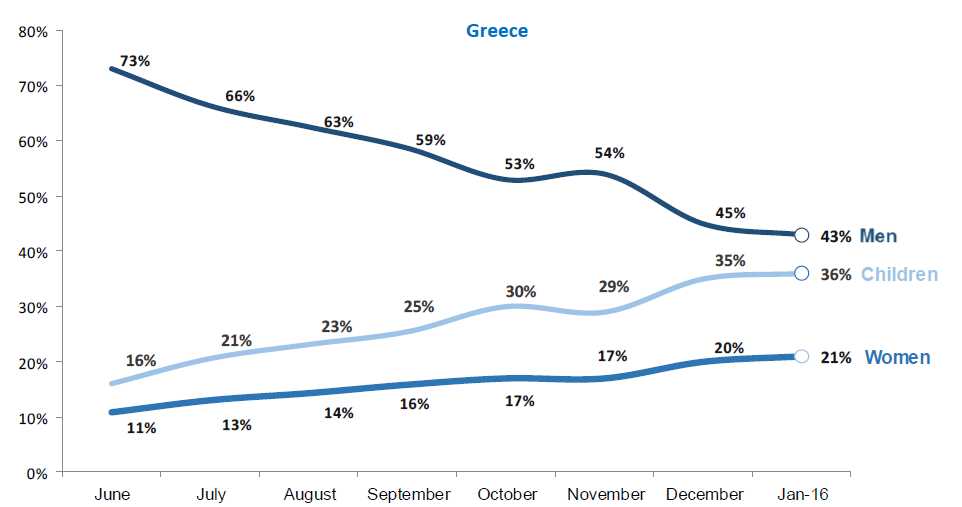}}
\par\end{centering}

\caption{\label{fig:Refugee-population-demogr}Refugee population demographics,
mid-2015 and early 2016 (Greece).}
\end{figure}

In this specific context, there were two additional factors that made
rapid response an imperative necessity. First, the winter season,
with rough sea condition and low temperatures, shrinking the survival
time in case of shipwrecks, since people did not have any protective
gear (thermal suits) other than life vests, if any. Second, as Figure
\ref{fig:Refugee-population-demogr} shows, the demographics of the
refugees arriving during the second half of 2015 shifted from primarily
men (73\%, June 2015) to primarily women and children (57\%, Jan.
2016) \cite{UNHCR-portal-url}. This means that the physical endurance
and the average survival time of people involved in shipwrecks were
becoming inherently worse, despite of weather conditions. 

It is clear that, despite any efforts to find solutions in the political
level for the refugee crisis at hand, the problems \emph{on the ground}
require well-informed decisions, high mobility and rapid response,
in order to save lives. The primary concern for the SSAR resources,
the medical teams, the volunteers and the NGOs assisting in the humanitarian
relief, as well as the proper logistics and warehouse management in
the first-reception islands, are all centered around the influx of
refugees via unsafe boats. The problem is inherently one of \emph{humanitarian
crisis management}; on the other hand, the major difficulty is not
the lack of civil infrastructure (e.g. electricity, open roads, communications,
etc) as in a large earthquake or a flood, but rather in the \emph{ability
to allocate adequate resources rapidly in various spots}. Therefore,
one of the most important and challenging tasks for a successful operation
in this context is to enlarge the time frame for short-term planning
deployment, i.e., improve the capabilities of \emph{early warning
\& prompt alerting}. It is a concept that is already included in emergency
planning and emergency operations in other contexts, for example in
forecasting water levels to issue early warnings for possible floods,
assessing weather conditions (humidity, temperature) to issue alert
warnings for possible wildfires in forests, etc. 

This study is an attempt to quantify and analyze in a systemic way
the task of developing such early warning/alert systems in the context
of refugee influx, using Greece and the Aegean Sea islands as the
main paradigm. To our knowledge, this is the first-ever attempt to
design such a system, since this refugee crisis itself and its geographical
properties is unique in its own way. The goal is to identify the underlying
statistical properties and the inherent ``system'' that produces
this influx, without any prior knowledge or insight of how the smuggling
networks operate near the Turkish coasts. Based on reliable data,
these models can then be used as guidelines for short-term forecasting
of the influx intensity, hence produce an invaluable operational asset
for planners, decision-makers and first-responders.

\section{Material and Data Overview}

\subsection{Daily arrivals data series}

This study is based on official data provided by UNHCR sources for
the daily arrivals of refugees in the Greek islands of the eastern
Aegean Sea \cite{UNHCR-data1-url,UNHCR-data2-url,UNHCR-data3-url}.
Specifically, UNHCR provides detailed daily logs of people registered
in the ``hotspot'' camps in the islands, as well as from verified
sources (other NGOs, Hellenic Coast Guard, Frontex). The reason for
having ``estimated'' and not exact numbers is that new arrivals
may not be registered immediately or UNHCR may not get informed promptly
by the other actors. The result of this data collection process is
a mashup of data for the daily arrivals in each one of the main reception
islands, a grand total for Greece and a series of explanatory reports.

\begin{figure*}[tbph]
\begin{centering}
\textsf{\includegraphics[width=17cm]{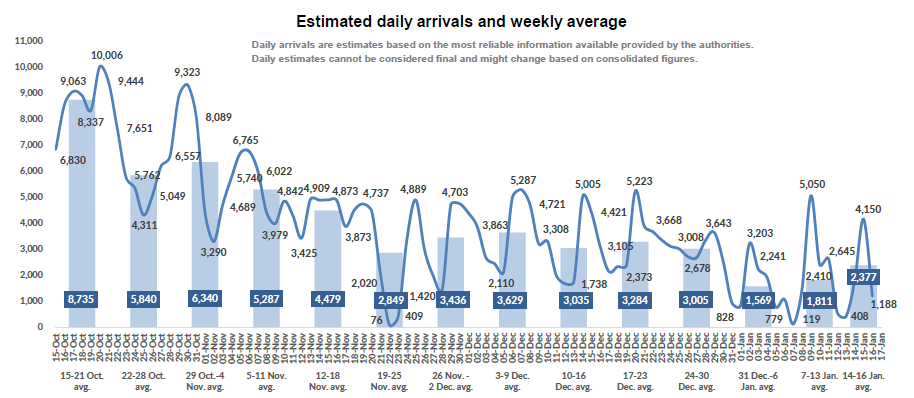}}
\par\end{centering}

\caption{\label{fig:Daily-arrivals-Greece}Estimated daily arrivals and weekly
averages for the entire Greece (Oct.2015-Jan.2016).}
\end{figure*}

\begin{figure}[htbp]
\begin{centering}
\textsf{\includegraphics[width=8.5cm]{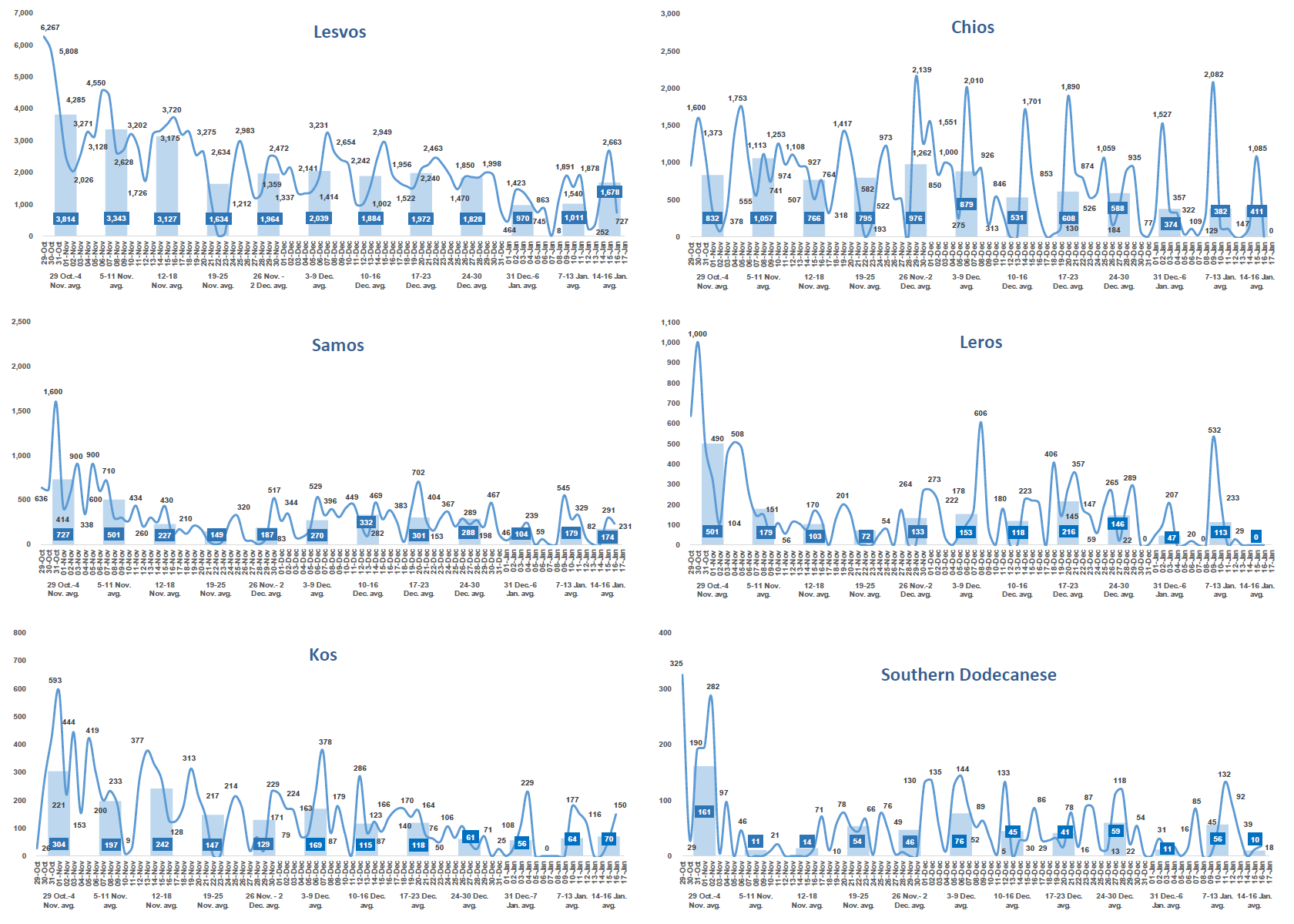}}
\par\end{centering}

\caption{\label{fig:Daily-arrivals-islands}Estimated daily arrivals and weekly
averages for the six main regions of interest in the Aegean Sea.}
\end{figure}

There are six main regions of interest in the Aegean Sea: the islands
of Lesvos, Chios, Samos, Leros, Kos and the rest of the southern Dodecanese
islands. Figure \ref{fig:Daily-arrivals-Greece} illustrates the main
data series for the composite total of daily arrivals in the entire
Aegean Sea, while Figure \ref{fig:Daily-arrivals-islands} illustrates
the individual daily arrivals in each of the six regions of interest.

In this study only the main data series was used for the analysis,
since more than 80\% of the influx is related mostly to Lesvos and
Chios. Furthermore, using the grand total of influx is inherently
more robust in terms of canceling individual noisy factors and enabling
the identification of global statistical trends. The time frame used
in this case for the data series is from 1-Oct-2015 to 16-Jan-2016,
a total span of 108 consecutive days (almost 15 weeks). The entire
data series is used in several analysis methods below, while a weekly-grouped
version of a slightly truncated data series in ``matrix'' mode (15x7
= 105 days) is used by other methods, as described in each case later
on.

\subsection{Software packages and hardware}

The main software packages that were used in this study were:
\begin{itemize}
\item Mathworks MATLAB v8.6 (R2015b), including: Signal Processing Toolbox,
System Identification Toolbox, Statistics \& Machine Learning Toolbox
\cite{Matlab-url}.
\item Additional toolboxes for MATLAB (own \& third-party) for specific
algorithms, as referenced later on in the corresponding sections.
\item WEKA v3.7.13 (x64). Open-Source Machine Learning Suite \cite{Weka-url}.
\item Spreadsheet applications: Microsoft Excel 2007, LibreOffice Sheet
5.1 (x64).
\item Custom-built programming tools in Java and C/C++ for data manipulation
(import/export).
\end{itemize}
The data experiments and processing were conducted using: (a) Intel
i7 quad-core @ 2.0 GHz / 8 GB RAM / MS-Windows 8.1 (x64), and (b)
Intel Atom N270 dual-core @ 1.6 GHz / 2 GB RAM / Ubuntu Linux 14.04
LTS (x32).

\section{Basic statistics}

\begin{figure}[htbp]
\begin{centering}
\textsf{\includegraphics[width=8.5cm]{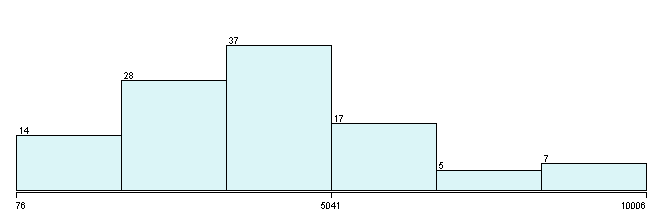}}
\par\end{centering}

\caption{\label{fig:Stats-Histogram-plot}Histogram plot of the daily arrivals
(108 points).}
\end{figure}

The standard histogram plot of the signal can provide the basic statistical
properties of the data when no time dependency (sequencing) is taken
into account. Figure \ref{fig:Stats-Histogram-plot} illustrates the
distribution of the daily arrivals (volume) for six bins, while Table
\ref{tab:Stats-Basic} contains the range statistics and the first
moments of the data \cite{Schaum-ProbStat-2009,Schaum-MathTabl-2012,Theodoridis.Konstantinos2009}.

\begin{table}[htbp]
\caption{\label{tab:Stats-Basic}Basic statistics of the daily arrivals (108
points).}

\centering{}%
\begin{tabular}{|c|c|}
\hline 
Parameter & Value\tabularnewline
\hline 
\hline 
minimum & 76\tabularnewline
\hline 
maximum & 10,006\tabularnewline
\hline 
median & 4,077\tabularnewline
\hline 
mean & 4,151.51\tabularnewline
\hline 
stdev & 2,216.79\tabularnewline
\hline 
skewness & 0.497\tabularnewline
\hline 
kurtosis & 0.081\tabularnewline
\hline 
\end{tabular}
\end{table}

From the histogram in Figure \ref{fig:Stats-Histogram-plot}, it is
evident that the left and the right section of the Gaussian-like distribution
around the mean are somewhat different, with the first (lower values)
being ``thicker'' than the standard deviation and the second (higher
values) being ``thinner'' but more widely spread in terms of maximum
range. This means that the values below the average daily arrivals
are somewhat more common, but the values above the average spread
to a higher range. In practice, the high influx rates in daily arrivals
are somewhat fewer, with larger absolute deviation against the mean,
compared to the corresponding low influx rates that are a bit more
common, with smaller deviation against the mean. This observation
is consistent with the daily reports from the people involved in the
actual registration process in the landing areas. 

In terms of statistics, these asymmetries are quantified by the kurtosis
parameter, which describes the ``sharpness'' of the central Gaussian-like
lobe of the histogram, and the skewness parameter, which describes
the asymmetry of the lobe against the mean value (for Normal distribution,
both kurtosis and skewness are zero). Here, kurtosis value is close
to zero, but the large positive value of skewness translates to a
``heavy right tail'' in the corresponding distribution -- roughly
11.1\% of the daily arrivals are above 6,700. This asymmetry is also
encoded by the clear difference between the median and the mean values,
where the second is larger (to the ``right'') but both below the
absolute middle of the value range, i.e., min+(max-min)/2=5,041.

One more important note is related to the value of standard deviation
$\left(\sigma\right)$: In any statistical distribution that can be
modeled effectively by Gaussian-like approximation, the range $\left[-\sigma\ldots+\sigma\right]$
contains roughly 68\% and $\left[-2\sigma\ldots+2\sigma\right]$ contains
roughly 95\% of the data \cite{Schaum-ProbStat-2009,Schaum-MathTabl-2012}.
This general observation is very important for translating the standard
deviation into a usable predictive factor, since in this case it means
that at least 2/3 of the daily arrivals are in the approximate range
of $\mu=4,151\pm2,217$. However, daily arrivals is a time series,
i.e., the data have specific time structuring (sequencing) and therefore
general predictive analytics, such as confidence ranges for the mean
value, are less important here compared to other auto-regressive and
decomposition approaches that are described later on. 

For completeness, the histogram in Figure \ref{fig:Stats-Histogram-plot}
was approximated by best-fit Poisson and Generalized Extreme Value
(GEV) \cite{GenExtrValDistr_2000} distributions. The data series
is clearly a zero-bounded set, compatible to the GEV formulation,
and it corresponds to ``events'' or ``arrivals'', compatible to
the Poisson formulation. The best-fit parameters of the distributions
in both cases were calculated by Maximum Likelihood Estimation (MLE)
for significance level 95\% $\left(\alpha=0.05\right)$. For Poisson,
parameter \emph{lambda} is $\lambda=4,151.5\pm12.5$, which is identical
to the Gaussian mean value. For GEV, parameters are \emph{shape} $\xi=-0.15\pm0.15$,
\emph{scale} $\sigma=1,980\pm297$ and \emph{location} $\mu=3,240\pm419$,
which makes it narrower than the Gaussian distribution and shifted
to the left, i.e., towards the lower bound (zero), as expected.

\begin{figure}[htbp]
\begin{centering}
\textsf{\includegraphics[width=8.5cm]{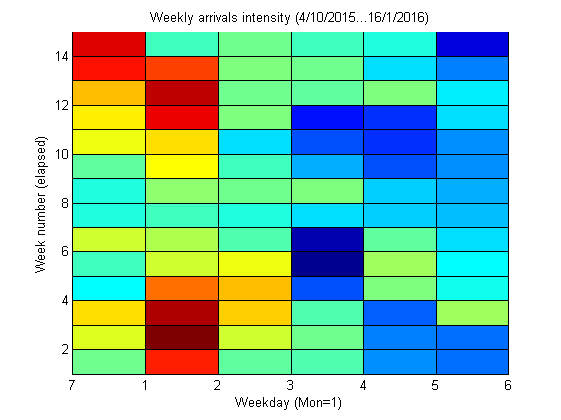}}
\par\end{centering}

\caption{\label{fig:Daily-arrivals-weekly}Daily arrivals grouped by week,
illustrating days of various influx intensity, from low (blue) to
high (red).}
\end{figure}

\begin{figure}[htbp]
\begin{centering}
\textsf{\includegraphics[width=8.5cm]{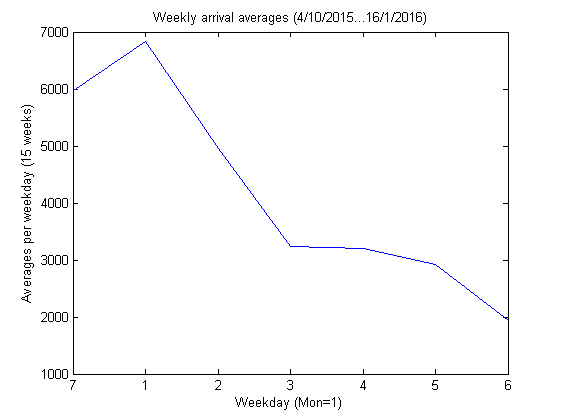}}
\par\end{centering}

\caption{\label{fig:Weekday-averages}Weekday averages of daily arrivals.}
\end{figure}

Figure \ref{fig:Daily-arrivals-weekly} illustrates the daily arrivals
grouped by week, starting from day 4 to day 108, producing 15x7 or
105 days in total, i.e., excluding the first three days of the original
data series. Cells are colored according to their relative influx
intensity, from low (blue) to high (red) and intermediate values (green/cyan).
The horizontal axis represents weekdays (Monday=1, Sunday=7) and the
vertical axis represents elapsed weeks from the start. Figure \ref{fig:Weekday-averages}
shows the corresponding weekday averages, taken against the columns
of the ``matrix'' illustrated by Figure \ref{fig:Daily-arrivals-weekly}.

The plots in Figure \ref{fig:Daily-arrivals-weekly} and Figure \ref{fig:Weekday-averages}
demonstrate a clear difference in the influx intensity between various
weeks and between various weekdays. Several external factors are associated
to specific dates in this time frame, e.g. days of calm weather or
political announcements related to the refugee crisis in Europe, which
explain such differences at some level. In any case, it is clear that
the first four weeks (early/mid-October 2015) and the last five weeks
(mid-December 2015, early/mid January 2016) are characterized by the
highest influx rates, with a relatively consistent preference in the
Sunday/Monday 48-hour time frame.

\section{Regression modeling}

Two variants of the typical regression analysis are applied to the
data series, in order to identify linear and periodic trends by predictive
modeling.

\subsection{Linear auto-regression}

The \emph{linear regression} approach \cite{Theodoridis.Konstantinos2009,Han-ARMA-1980}
is the most common and most basic model to formulate statistical dependencies
between two series of data, usually an ``input'', i.e., an external
factor or control variable, and an ``output'', i.e., an associated
result. The most common formulation is:

\begin{equation}
\hat{y}\left(t\right)=\overrightarrow{b_{k}}\cdot\overrightarrow{u_{k}}\left(t\right)+e\left(t\right)\label{eq:linear-autoregr}
\end{equation}
where $\overrightarrow{u_{k}}\left(t\right)$ is the input vector
of size $k$ at time step $t$, $\overrightarrow{b_{k}}$ is the (static)
vector of regression coefficients, $\hat{y}\left(t\right)$ is the
estimated output and $e\left(t\right)$ is the model error at time
step $t$, i.e., $e\left(t\right)=\left\Vert \hat{y}\left(t\right)-y\left(t\right)\right\Vert ^{2}$.
When there is no evident input and $y\left(t\right)$ is the data
series of the output generated by an unknown process, then the input
vector corresponds to some of the previous output values, typically
a fixed-length window:

\begin{equation}
\overrightarrow{u_{k}}\left(t\right)\triangleq\overrightarrow{u_{n}}\left(t\right)=\left\langle y\left(t\right),y\left(t-1\right),\ldots,y\left(t-n+1\right)\right\rangle \label{eq:linear-autoregr-inputvec}
\end{equation}
where $\overrightarrow{u_{n}}\left(t\right)$ is the vector that consists
of the current plus $n-1$ previous outputs.

In this study, a linear auto-regression model as in Eq.\ref{eq:linear-autoregr}
was introduced to approximate the daily arrivals, using an auto-regressive
window as in Eq.\ref{eq:linear-autoregr-inputvec} with fixed length
$n=13$ days as the input. The best-fit model was:

\begin{equation}
\begin{array}{cc}
\hat{y}_{n}\backsimeq & 1294.1+1.168\cdot y_{n-1}-0.458\cdot y_{n-2}\\
 & +0.315\cdot y_{n-3}-0.287\cdot y_{n-4}\\
 & +0.144\cdot y_{n-10}+0.155\cdot y_{n-11}+\ldots
\end{array}\label{eq:linear-autoregr-solution}
\end{equation}
with Mean Absolute Prediction Error (MAPE): $err_{MAPE}=E\left[\left|e\left(t\right)\right|\right]=771.3$
and Root Mean Squared Error (RMSE): $err_{RMSE}=\sqrt{E\left[e\left(t\right)^{2}\right]}=1009.2$
\cite{Theodoridis.Konstantinos2009}, where the errors represent true
output values (daily arrivals). 

In this model, the most interesting factor is not the prediction accuracy
per se, but the identification of the largest regression coefficients.
Specifically, the result in Eq.\ref{eq:linear-autoregr-solution}
shows that the spot value of the daily arrivals at any time step is
associated primarily with the spot values of the four previous days
and the values 10 and 11 days before, i.e., \emph{not} so much by
the values of the time frame 5-9 days before. This finding is a hint
of a possible periodic behavior in the data series that needs further
investigation with appropriate non-linear (periodic) components in
the regression model. The following section enhances the model of
Eq.\ref{eq:linear-autoregr} with such factors and identifies the
predominant periodic trends in the data.

\subsection{Cosine-linear regression}

In order to identify the periodicity of the data series of daily arrivals,
a cosine term is introduced in the model of Eq.\ref{eq:linear-autoregr},
resulting in the new \emph{cosine-linear regression} model: 
\begin{equation}
\hat{y}\left(t\right)=\left(a\cdot\cos\left(b\cdot t+c\right)\right)+\left(d\cdot t+c_{0}\right)\label{eq:cosine-linear-regr}
\end{equation}
where $d$ and $c_{0}$ define a standard linear component as in Eq.\ref{eq:linear-autoregr},
whereas $a$, $b$ and $c$ define the cosine component. In particular,
$a$ is the scaling parameter, $b$ is the periodic parameter and
$c$ is the corresponding phase. Here, the time sequencing parameter
$t$ is the only input introduced in the model, in contrast to Eq.\ref{eq:linear-autoregr}
where previous values of the data series itself were used as input
(hence the term \emph{auto}-regression).

\begin{figure}[htbp]
\begin{centering}
\textsf{\includegraphics[width=8.5cm]{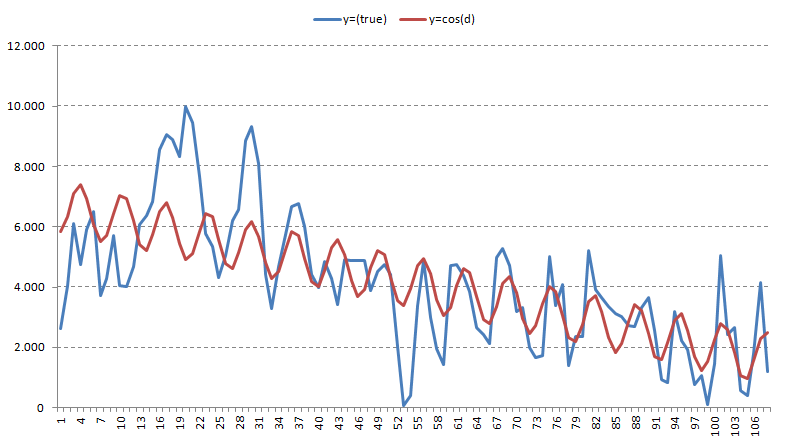}}
\par\end{centering}

\caption{\label{fig:cosine-linear-autoregr}Comparison of true and predicted
daily arrivals, using the best-fit (SSE) cosine-linear regression
model of Eq.\ref{eq:cosine-linear-regr-solution}.}
\end{figure}

Using the cosine-linear regression model of Eq.\ref{fig:cosine-linear-autoregr},
the best-fit parameters where calculated as follows:
\begin{equation}
\hat{y}\left(t\right)=\left(875\cdot\cos\left(0.97\cdot t-2.85\right)\right)+\left(-47\cdot t+6669\right)\label{eq:cosine-linear-regr-solution}
\end{equation}
that is: $a\backsimeq875.2$, $b\backsimeq0.968$, $c\backsimeq-2.851$,
$d\backsimeq-47.065$ and $c_{0}\backsimeq6669.5$, optimal in the
sense of Sum of Squared Errors (SSE) \cite{Theodoridis.Konstantinos2009}.

As with the simple linear auto-regression of Eq.\ref{eq:linear-autoregr-solution},
the most interesting factor here is the values of the parameters,
rather that the absolute prediction error. In particular, the predominant
periodic trend, i.e., the major ``period'' of the daily arrivals,
can be calculated from the periodic parameter $b$ as: $T_{C}=\nicefrac{2\pi}{b}\backsimeq6.5$
(days). The linear trend or ``slope'' of the data series is the
parameter $d=-47$, which is ``downward''. This translates to 47
fewer arrivals each day, i.e., roughly equivalent to just one boat
less. Figure \ref{fig:cosine-linear-autoregr} illustrates the true
and the best-fit cosine-linear regression model as described in Eq.\ref{eq:cosine-linear-regr-solution}.

\section{Spectral analysis}

One of the most common approaches in analyzing the periodic properties
of a signal is spectral decomposition via the \emph{Fourier transformation},
specifically the \emph{Discrete Fourier Transform} (DFT) or most commonly
the \emph{Fast Fourier Transform} (FFT) for discrete-valued signals
\cite{Opph-DSP-1975,Proa-DSP-1989,Ther-RndSig-1992,Harris_tukey_1978,Hsu-SigSys-1995,Opph-SigSys-1983}.
It is the most popular way of approximating a data sequence, structured
in the time domain, by a series of sine and cosine components, structured
in the frequency domain. In this way, the same signal remains the
same but its representation is translated into frequency components,
making its spectral properties clear and detailed.

The DFT is defined as: 
\begin{equation}
Y_{k}\triangleq\sum_{n=0}^{N-1}y_{n}\cdot e^{-i2\pi k\frac{n}{N}}\;,\;k\epsilon\mathbf{Z}\label{eq:DFT-complex}
\end{equation}
or, in analytical form:
\begin{equation}
Y_{k}\triangleq\sum_{n=0}^{N-1}y_{n}\cdot\left(\cos\left(-2\pi k\frac{n}{N}\right)+i\cdot\sin\left(-2\pi k\frac{n}{N}\right)\right)\;,\;k\epsilon\mathbf{Z}\label{eq:DFT-analytical}
\end{equation}
where $y_{n}$ are the signal samples, $N$ is the size of the data
series, $k$ is each frequency under consideration and $Y_{k}$ is
the corresponding (complex) frequency component. Under DFT, the signal
is considered periodic ($T=N$) and, due to the Nyquist-Shannon sampling
theorem \cite{Opph-DSP-1975,Proa-DSP-1989}, the (discrete) spectrum
is ``mirrored'' around $\pi,$ hence the maximum identifiable frequency
here is $\nicefrac{N}{2}$, which correspond to ``changes per period''. 

\begin{figure}[htbp]
\begin{centering}
\textsf{\includegraphics[width=8.5cm]{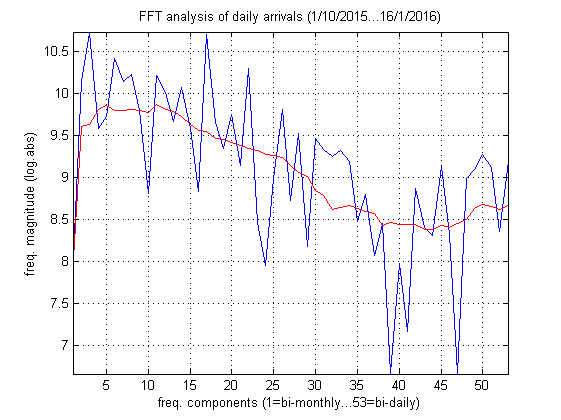}}
\par\end{centering}

\caption{\label{fig:FFT-logplot}Spectral representation (FFT log-plot) of
the complete data series of daily arrivals.}
\end{figure}

Figure \ref{fig:FFT-logplot} illustrates the spectral representation
(FFT) of the complete data series of daily arrivals, using full resolution
$\left(N=108\right)$. The blue line is the half-spectrum log-plot,
i.e., the rightmost point (max) on the horizontal axis corresponds
to $\omega=\pi$, while the red line is the corresponding moving average
with window size 20. 

The spectral components in Figure \ref{fig:FFT-logplot}, represented
by each point on the horizontal axis, correspond to the full range
of frequencies analyzed, from $f_{min}=\frac{1}{N}=\frac{1}{108}$
to $f_{max}=\frac{N/2}{N}=\frac{1}{2}$. This frequency range represents
``changes'' and it runs from one such ``change'' for the entire
data series available ($T_{max}=N=108$ days) to the maximum resolution,
which is half the entire size ($T_{min}=2$ days), according to Nyquist
theorem \cite{Opph-DSP-1975,Proa-DSP-1989,Harris_tukey_1978}. Since
the vertical axis is log-scaled, every +1 in value corresponds to
10 times larger energy in the specific component. 

The size of the data series is relatively small, hence the spectral
analysis in Figure \ref{fig:FFT-logplot} is useful for a qualitative,
rather than quantitative assessment of the signal. However, the moving
average highlights some important aspects, already identified by the
preliminary analysis via linear and cosine-linear regression. Specifically,
the power density profile illustrates a typical low-frequency signal,
with most of its energy packed in the lower 1/3 of the FFT spectrum.
The point where the moving average (red line) crosses downwards to
energy magnitudes lower than 9.5 is roughly at $x_{L}=17.5$ on the
horizontal axis; since it is scaled from $\left\{ f_{min}:\:x_{min}=1\right\} $
to $\left\{ f_{max}:\:x_{max}=\left\lceil \frac{108}{2}\right\rceil =54\right\} $
as described earlier, this point corresponds roughly to the rescaled
point $f_{L}$ according to:

\begin{equation}
\begin{array}{cc}
f_{L} & =\left(\frac{x_{L}-x_{min}}{x_{max}-x_{min}}\right)\cdot\left(f_{max}-f_{min}\right)+f_{min}\\
 & =\left(\frac{17.5-1}{54-1}\right)\cdot\left(\frac{1}{2}-\frac{1}{108}\right)+\frac{1}{108}\\
 & \simeq0.162
\end{array}\label{eq:FFT-logplot-rescaling}
\end{equation}

In other words, the daily arrivals signal is (for the most part) bounded
by the upper frequency $f_{L}=0.162$ or, in terms of period, $T_{L}=\frac{1}{f_{L}}\simeq6.1714$
(days). This limit is very close to the period $T_{C}=6.5$ identified
by the cosine-linear regression model earlier, which strongly suggests
that the data series is indeed periodic with a major period of almost
(less than) a full week.

\section{Auto-Regressive Moving-Average}

The statistical and frequency properties of the daily arrivals data
series was analyzed via pairwise correlation, phase diagrams and full
system identification, specifically by Auto-Regressive Moving Average
(ARMA) approximations, as described below.

\subsection{Auto-correlation \& phase}

Pairwise correlation produces a quantitative metric for the statistical
dependencies between values of two data series at different lags.
In the case when a single data series is compared to itself, the \emph{auto-correlation}
corresponds to the statistical dependencies between subsequent values
of the same series \cite{Hsu-SigSys-1995,Hamming-filters-1989}. Hence,
value pairs with high correlation correspond to regular patterns in
the series, i.e., encode periodicity at smaller or larger scales,
according to the selected lag.

\begin{figure}[htbp]
\begin{centering}
\textsf{\includegraphics[width=8.5cm]{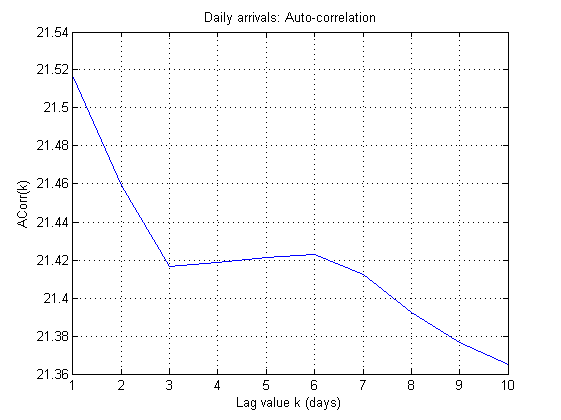}}
\par\end{centering}

\caption{\label{fig:Auto-correlation-plot}Auto-correlation plot of the daily
arrivals for lag limit $k=\pm10$. }
\end{figure}

In this study, the daily arrivals were analyzed via auto-correlation
with a lag limit of $k\pm10$ against the current day. This limit
was selected as appropriate after the preliminary analysis via regression
modeling that confirmed strong periodicity and frequency components
below the 7 days boundary (see above). Figure \ref{fig:Auto-correlation-plot}
presents the plot of the auto-correlation vector of the entire daily
arrivals data series; the auto-correlation vector is symmetric around
the central point ($k=0$), hence only the positive half-width plot
($1\leq k\leq10$) is included here.

A typical chaotic or semi-stochastic signal is expected to exhibit
an exponentially decreasing profile in its auto-correlation plot:
sharper ``drop'' of the profile as the lag value $k$ increases
means smaller window of dependency between subsequent values, i.e.,
a higher-frequency signal. In contrast, lag values that present a
curve higher than the expected asymptotically vanishing profile correspond
to significant statistical dependencies at this scale, i.e., strong
periodic trends.

As in the case of regression modeling, the auto-correlation plot in
Figure \ref{fig:Auto-correlation-plot} reveals strong periodic trends
around the 6-days threshold. It also reveals strong low-frequency
energy profile, since it exhibits a distinct high peak at $k=1$ and
exponentially decreasing gradually to $k=3$; this means that daily
arrivals are strongly related to previous values of up to three days.
The plot also shows that this strong dependencies remain valid for
at least six days in total. 

\begin{figure}[htbp]
\begin{centering}
\textsf{\includegraphics[width=8.5cm]{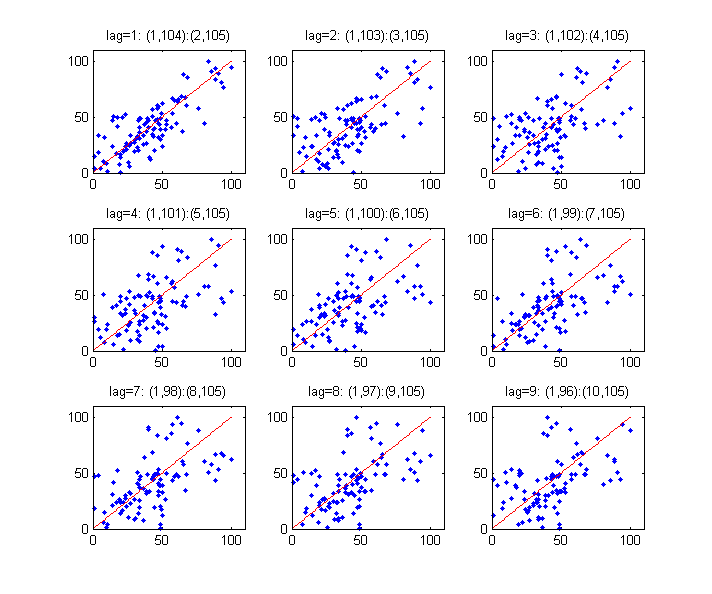}}
\par\end{centering}

\caption{\label{fig:Phase-plots}Phase diagrams of the daily arrivals for lag
limit $k=\pm9$.}
\end{figure}

Figure \ref{fig:Phase-plots} presents the \emph{phase diagrams} of
the entire daily arrivals data series, i.e., the 2-D plot of subsequent
values $y\left(t\right)$ against $y\left(t+k\right)$, scaled down
by 100 and separated by different lag values of up to nine days ($1\leq k\leq9$).
Each pair is presented by a dot (blue), while the diagonal line (red)
corresponds to the symmetric boundary of $k=0$.

In accordance to the auto-correlation plots, strong statistical dependencies
between value pairs in the signal appear as clusters; the closer they
are to the symmetric boundary, the smaller is the lag separation between
similar values. In other words, a low-frequency signal appears with
most value pairs ``packed'' around the symmetric boundary. The exact
frequency components of the signal (spectral profile) affect how this
distribution ``spreads'' wider around the symmetric boundary as
the lag value increases, i.e., the separation between the values of
each pair.

Although the phase plots in Figure \ref{fig:Phase-plots} are not
conclusive in a similar way as with the auto-correlation plot in Figure
\ref{fig:Auto-correlation-plot}, it is evident that the plot for
a one-day lag ($k=1$) is clearly more ``packed'' around the symmetric
boundary than in any other lag value. In accordance to the basic statistics
presented earlier in Figure \ref{fig:Stats-Histogram-plot} and Table
\ref{tab:Stats-Basic}, the clusters in all plots appear more dense
below the value 60 (i.e., less than 6,000 arrivals a day). Furthermore,
within that range, a somewhat tighter ``packing'', similar to the
one for $k=1$, reappears when $5<k<8$; again, evidence that there
is some periodic trend within that range (lag in days).

\subsection{ARMA system identification}

More generic and powerful than auto-correlation or linear regression
alone, the \emph{Auto-Regressive Moving Average} (ARMA) model is the
standard approach for describing any linear digital filter or signal
generator in the time domain. It is essentially a combination of an
auto-correlation component that relates the current outputs to previous
ones and a smoothing component than averages the inputs over a fixed-size
window.

The typical linear ARMA model is described as \cite{Hamming-filters-1989,Porat-signals-1994,Ther-RndSig-1992,Han-ARMA-1980}:

\begin{equation}
A_{m}\left(z\right)\ast\overrightarrow{y}\left(t\right)=B_{k}\left(z\right)\ast\overrightarrow{u}\left(t\right)+e\left(t\right)\label{eq:ARMA-generic}
\end{equation}
where $\overrightarrow{u_{k}}\left(t\right)$ is the input vector
of size $k$ at time step $t$, $\overrightarrow{y}\left(t\right)$
is the output vector of size $m$ (i.e., the current plus the $m-1$
previous ones), $B_{k}\left(z\right)$ is the convolution kernel for
the inputs, $A_{m}\left(z\right)$ is the convolution kernel for the
outputs and $e\left(t\right)$ is the residual model error. Normally,
$A_{m}\left(z\right)$ and $B_{k}\left(z\right)$ are vectors of scalar
coefficients that can be fixed, if the model is static, or variable,
if the model is adaptive (constantly ``retrained''). 

Both coefficient vectors, as well as their sizes, are subject to optimization
of the model design according to some criterion, which typically is
the minimization of the residual error $e\left(t\right)$. In practice,
this is defined as $e\left(t\right)=\left\Vert \hat{y}\left(t\right)-y\left(t\right)\right\Vert ^{2}$,
where $\hat{y}\left(t\right)$ is the ARMA-approximated output and
$y\left(t\right)$ is the true (measured) process output. The sizes
$m$ and $k$ are the \emph{orders} of the model and they are usually
estimated either by information-theoretic algorithms \cite{Hamming-filters-1989,Porat-signals-1994,Han-ARMA-1980}
or by exploiting known properties (if any) of the generating process,
e.g. with regard to its periodicity. Such a model is described as
ARMA($m$,$k$), where AR($m$) is the auto-regressive component and
MA($k$) is the moving-average component. 

In approximation form, expanding the convolutions and estimating the
current output $\hat{y}\left(t\right)$, the analytical form of Eq.\ref{eq:ARMA-generic}
is:

\begin{equation}
\hat{y}\left(t\right)=\sum_{i=1}^{m}\left(a_{i}\cdot y\left(t-i\right)\right)+\sum_{j=0}^{k}\left(b_{j}\cdot x\left(t-j\right)\right)+e\left(t\right)\label{eq:ARMA-analytical}
\end{equation}

The error term $e\left(t\right)$ in Eq.\ref{eq:ARMA-analytical}
can also be expanded to multiple terms of a separate convolution kernel,
similarly to $A_{m}\left(z\right)$ and $B_{k}\left(z\right)$, but
it is most commonly grouped into one scalar factor, i.e., with an
order of one. In such cases, the model and be described as ARMA($m$,$k$,$q$)
where $q>1$ is the order of the convolution kernel for the error
term.

When applied to a signal generated by a process of unknown statistical
properties, an ARMA approximation of it reveals a variety of important
properties regarding this process. In practice, the (estimated) order
$m$ of the AR component shows how strong the statistical coupling
is between subsequent outputs, while the order $k$ of the MA component
shows the ``memory'' of the process, i.e., how far in the past inputs
the process ``sees'' in order to produce the current output.

In the current study, preliminary analysis via auto-correlation, linear
regression and cosine-linear regression (see previous sections) has
revealed strong periodic components that can be exploited here. Additionally,
the ARMA models were designed and trained using arbitrary ranges for
their orders to verify and optimize the initial choices.

Several ARMA models where designed and optimized for approximating
the daily arrivals data series for exploring the necessary orders
and the distribution of magnitude in the corresponding coefficient
vectors. Since the daily arrivals is inherently an auto-regressive
process, with each spot value depending heavily on the values of the
previous days, the tested orders for the AR where bounded between
1 (just the previous day) and 21 (three full weeks back); the weekday
was used as the input with MA order of 1 (no averaging over repeating
weekday cycles); and the error term was modeled with convolution kernels
of orders up to 3.

Using an ARMA(9,1,3) model, the optimal approximation of the daily
arrivals yields:

\begin{equation}
\begin{array}{cc}
A_{9}\left(z\right) & =1-0.8887\cdot z^{-1}+0.1247\cdot z^{-2}+0.2971\cdot z^{-3}\\
 & -0.3747\cdot z^{-4}+0.1526\cdot z^{-5}-0.1265\cdot z^{-6}\\
 & -0.1357\cdot z^{-7}+0.164\cdot z^{-8}-0.144\cdot z^{-9}
\end{array}\label{eq:ARMA913-Avec}
\end{equation}
and:

\begin{equation}
B_{9}\left(z\right)=48.94\cdot z^{-3}\label{eq:ARMA913-Bvec}
\end{equation}

The $A_{9}\left(z\right)$ and $B_{9}\left(z\right)$ polynomials
are the (optimal) convolution kernels for the AR(9) and MA(1) components,
respectively. In both polynomials, $z^{-n}$ is the delay factor of
the kernel, as described by the analytical form of Eq.\ref{eq:ARMA-analytical}.
Hence, the coefficient of $z^{-n}$ in $A_{9}\left(z\right)$ is essentially
$a_{n}$, i.e., the magnitude for the auto-regressive factor (output)
$n$ days back. 

\begin{figure}[htbp]
\begin{centering}
\textsf{\includegraphics[width=8.5cm]{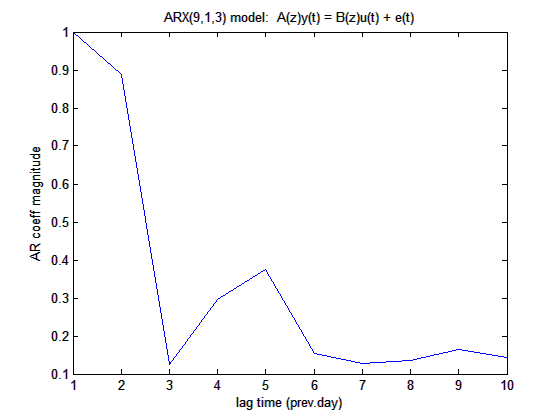}}
\par\end{centering}

\caption{\label{fig:ARMA913-Avec-plot}Magnitude (abs) plot of the convolution
kernel of Eq.\ref{eq:ARMA913-Avec}; horizontal axis is: $\left[1,A_{9}+1\right]$. }
\end{figure}

The most important results from Eq.\ref{eq:ARMA913-Avec} are: (a)
the $z^{-1}$ coefficient is the largest and very close to 1, signifying
strong low-frequency spectral components, and (b) the $z^{-3}$ and
$z^{-4}$ coefficients are at least twice as large as any other coefficient
in this AR(9) convolution kernel, signifying strong within-week periodic
trends in the daily arrivals. Figure \ref{fig:ARMA913-Avec-plot}
presents the AR(9) coefficients and illustrates the clear increase
in magnitude of these two coefficients at lags 4 and 5. 

While the ARMA(9,1,3) approximation is useful for evaluating the weekly
time frames, a more complex model was also applied for a better minimal-error
approximation of the daily arrivals. Specifically, an ARMA(21,1,1)
was estimated for a full three-week time frame for AR with singular
MA and error convolution kernels. In this case, the optimal model
design yields:

\begin{equation}
\begin{array}{cc}
A_{21}\left(z\right) & =1-1.608\cdot z^{-1}+0.9277\cdot z^{-2}+0.3233\cdot z^{-3}\\
 & -1.142\cdot z^{-4}+0.9955\cdot z^{-5}-0.4887\cdot z^{-6}\\
 & -0.1842\cdot z^{-7}+0.4908\cdot z^{-8}-0.4291\cdot z^{-9}\\
 & -0.02225\cdot z^{-10}+0.235\cdot z^{-11}-0.1331\cdot z^{-12}\\
 & -0.1298\cdot z^{-13}+0.1242\cdot z^{-14}-0.0892\cdot z^{-15}\\
 & +0.2391\cdot z^{-16}-0.3209\cdot z^{-17}+0.4167\cdot z^{-18}\\
 & -0.4449\cdot z^{-19}+0.1795\cdot z^{-20}+0.1159\cdot z^{-21}
\end{array}\label{eq:ARMA2111-Avec}
\end{equation}
and:

\begin{equation}
B_{21}\left(z\right)=57.98\cdot z^{-1}\label{eq:ARMA2111-Bvec}
\end{equation}

As in the case of the AR(9) component in Eq.\ref{eq:ARMA913-Avec},
this AR(21) component in Eq.\ref{eq:ARMA2111-Avec} reveals very useful
information about the generating process. More specifically, it is
clear that: (a) the magnitude of all coefficients fade asymptotically
as the lag increases, signifying a signal with strong low-frequency
spectral components, and (b) there is a clear pattern of alternating
larger and smaller magnitudes in the components, signifying periodic
trends within a window smaller than the order of AR(21).

\begin{figure}[htbp]
\begin{centering}
\textsf{\includegraphics[width=8.5cm]{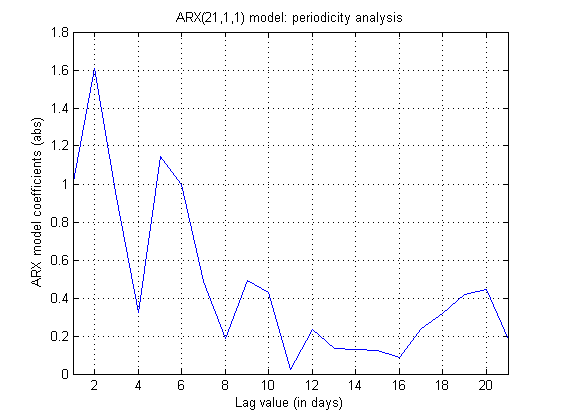}}
\par\end{centering}

\caption{\label{fig:ARMA2111-Avec-plot}Magnitude (abs) plot of the convolution
kernel of Eq.\ref{eq:ARMA2111-Avec}; horizontal axis is: $\left[1,A_{21}+1\right]$.}
\end{figure}

Figure \ref{fig:ARMA2111-Avec-plot} presents the AR(21) coefficients
and verifies these observations regarding the distribution of magnitudes.
In this case, the peaks are at lags 2, \{5,6\}, \{9,10\}, etc. There
is also evidence of longer-term dependencies beyond lag 18 (end of
3rd week in the time frame), although in many cases such patterns
may be related to noise or other approximation artifacts rather than
the generating process itself. 

\begin{figure}[htbp]
\begin{centering}
\textsf{\includegraphics[width=8.5cm]{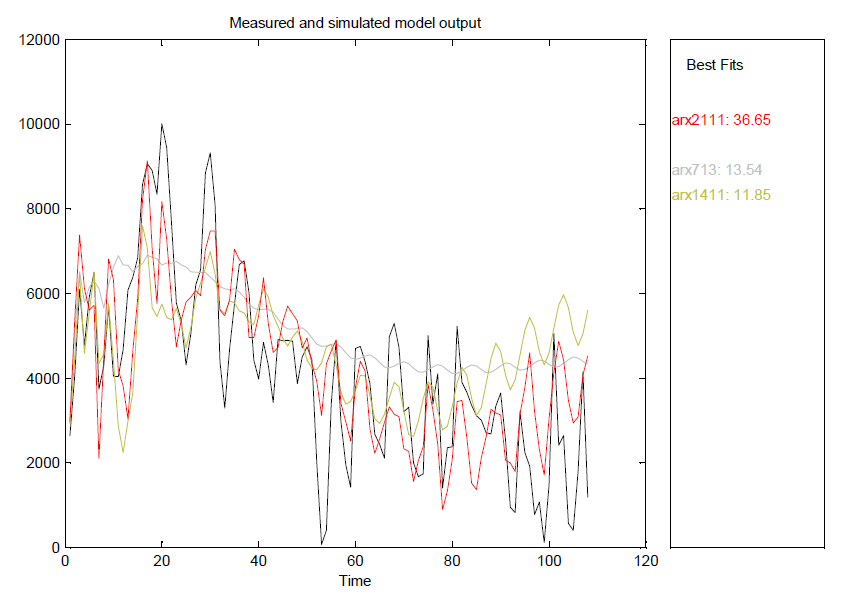}}
\par\end{centering}

\caption{\label{fig:ARMA-3approx-plots}Comparative plot of one-week ARMA(7,1,3)
(gray), two-week ARMA(14,1,1) (olive) and three-week ARMA(21,1,1)
(red) approximation models for the daily arrivals data series (black).}
\end{figure}

Figure \ref{fig:ARMA-3approx-plots} presents three ARMA model approximations
of the daily arrivals with different AR orders. Specifically, a one-week
ARMA(7,1,3) (gray), a two-week ARMA(14,1,1) (olive) and a three-week
ARMA(21,1,1) (red) is presented against the true data series of the
daily arrivals (black). It is clear that, as the AR order increases,
the approximation becomes better and more detailed than the general
trend. These results show that an AR(21), i.e., a three-week auto-regressive
model, can produce a practically usable formulation for the \emph{short-term
forecasting} of daily arrivals.

\section{Matrix factorization - Component analysis}

Spectral decomposition and frequency analysis of a signal often involves
some transformation to the spectral domain, as it was described earlier
for FFT with Eq.\ref{eq:DFT-complex} and Eq.\ref{eq:DFT-analytical}.
An alternative approach is to reformulate the original signal into
a matrix form and assume multiple signals of shorter size, in order
to analyze their similarities in the spectral domain. This is typically
conducted by some form of \emph{Matrix Factorization} (MF) that expresses
the original matrix as a product of two other matrices, the ``spectral
components'' and the corresponding coefficients. In the context of
system identification, MF can be used as a very powerful tool for
discovering the periodic trends and the spectral properties of the
original signal and, hence, the generating process in question.

In this study, MF was employed for spectral decomposition of the daily
arrivals in various forms, including SVD, PPCA, ICA and K-SVD, as
described below.

\subsection{SVD}

The \emph{Singular Value Decomposition} (SVD) \cite{Theodoridis.Konstantinos2009}
of a matrix is one of the most widely used algorithms in Linear Algebra
with regard to rank and dimensionality reduction. Given a $l\times n$
matrix $Y$ of rank $r\leq\min\left\{ l,n\right\} $, SVD transforms
it to a product as:
\begin{equation}
Y=U\cdot\left[\begin{array}{cc}
A^{\frac{1}{2}} & O\\
O & 0
\end{array}\right]\cdot V^{H}\label{eq:SVD-definition}
\end{equation}
where $A^{\frac{1}{2}}$ is the $r\times r$ diagonal matrix with
elements $\sqrt{\lambda_{k}}$, and $\lambda_{k}$ are the $r$ non-zero
eigenvalues of the associated matrix $Y^{H}\cdot Y$. In other words,
SVD transforms the original matrix $Y$ into a special diagonal form
that includes the eigenvalues, which provide a very useful description
of its ``spectral'' components included in the eigenvector matrices
$U$ and $V$. This approach is being widely used for decades in Pattern
Recognition for a variety of applications, from signal compression
and dimensionality reduction to feature generation and image coding
\cite{Theodoridis.Konstantinos2009,eladbook,allen_svd_inconsistency}.

\begin{figure}[htbp]
\begin{centering}
\textsf{\includegraphics[width=8.5cm]{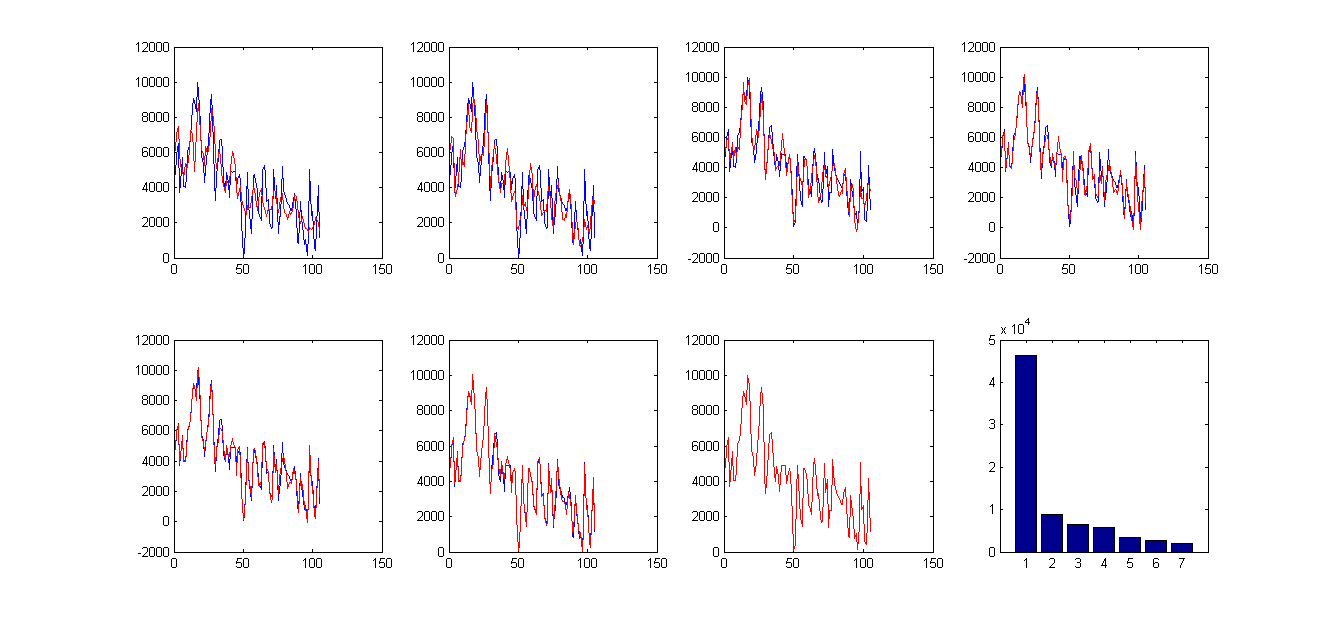}}
\par\end{centering}

\caption{\label{fig:SVD-approx-plot}SVD spectral approximation (red) of the
weekly-grouped arrivals, plotted against the true data series (blue);
the horizontal axis is the time (days); the histogram at the bottom-right
corner is the same components ranked according to their spectral energy.}
\end{figure}

In this study, the daily arrivals data series was restructured into
a weekly-grouped version of a slightly truncated version (15x7 = 105
days), as it is illustrated in Figure \ref{fig:Daily-arrivals-weekly}.
The purpose here is to decompose the data series into ``weekly trends''
via SVD, according to Eq.\ref{eq:SVD-definition}, and investigate
the significance of each individual component. Instead of examining
the eigenvalues in the corresponding matrix $A^{\frac{1}{2}}$, the
original data series is approximated by using 1 to 7 (all) SVD components,
in order to estimate their relative ``coding'' efficiency. Figure
\ref{fig:SVD-approx-plot} illustrates this approach, with the top-level
sub-plot referring to reconstruction by only the first (rank-1) SVD
component and subsequent sub-plots referring to reconstructions by
increasing number of components up to 7 (full-rank). The bottom-right
histogram illustrates the corresponding SVD components ranked according
to their spectral energy.

It is clear from the results in Figure \ref{fig:SVD-approx-plot}
that the daily arrivals is a signal with strong low-frequency signature.
The histogram, as well as the first row of sub-plots, illustrate that
the data series can be approximated effectively by the three or four
major spectral components. This means that if a predictive model is
required, the corresponding eigenvectors and eigenvalues calculated
by SVD can be used for rank-3 or rank-4 approximations, respectively,
for robust and noise-resilient forecasting.

\subsection{PPCA}

In Linear Algebra, \emph{Principal Component Analysis} (PCA) is a
MF algorithm for expressing a matrix as a result of \emph{orthogonal
projections}. In practice, it is a statistical procedure that uses
an \emph{orthogonal linear transformation} of a set of observations
into a set of values of linearly uncorrelated variables, called \emph{principal
components}. The PCA algorithm is formally known as \emph{Karhunen-Loeve
Transform} (KLT) \cite{Theodoridis.Konstantinos2009} and, like SVD,
it is being used for many years in Pattern Recognition for signal
coding, dimensionality reduction, feature generation, etc.

Formally, the KLT is defined as:
\begin{equation}
P=Y\cdot W\:\Leftrightarrow\:Y=P\cdot W^{H}\label{eq:KLT-definition}
\end{equation}
where $P$ is the ``loading vectors'' matrix, $Y$ is the original
data matrix and $W$ is the matrix whose columns are the eigenvectors
of $Y^{H}\cdot Y$. It is clear from Eq.\ref{eq:SVD-definition} and
Eq.\ref{eq:KLT-definition} that KLT is closely related to SVD, as
both of them involve the corresponding eigenvectors, i.e., they employ
orthogonal projections that are error-optimal in the mean-square-error
(MSE) sense. This means that, as with SVD (which is more general),
if only some of the ``spectral'' components are to be used for MSE-optimal
approximation of the original data, Eq.\ref{eq:KLT-definition} can
be used with the matrix $W$ truncated as to include only the first
$k$ eigenvectors (``basis'' columns). 

The \emph{Probabilistic Principal Component Analysis} (PPCA) \cite{Bishop_ppca_1999}
is an extension to PCA in the sense that it includes a parametric
probabilistic model for the signal (typically Gaussian, estimated
via Maximum Likelihood). This provides an efficient way to deal with
missing data, outliers and increased noise resiliency. The KLT formulation
is extended to include non-zero mean $\mu$ and residual error $\varepsilon$:
\begin{equation}
Y=P\cdot W^{H}+\mu+\varepsilon\label{eq:PPCA-definition}
\end{equation}

\begin{figure}[htbp]
\begin{centering}
\textsf{\includegraphics[width=8.5cm]{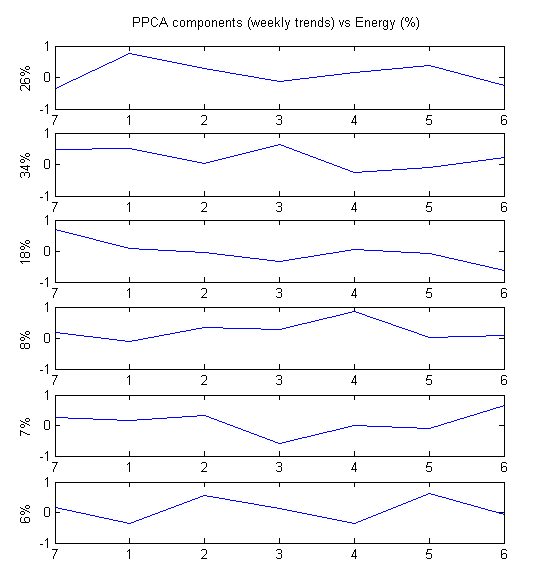}}
\par\end{centering}

\caption{\label{fig:PPCA-components-plot}PPCA components of the weekly-grouped
daily arrivals; horizontal axis is the weekdays; \% labels in the
left show the relative variance (energy) described by each component. }
\end{figure}

In this study, PPCA was used in a similar way as with SVD described
earlier, i.e, the daily arrivals data series was restructured into
a weekly-grouped version of a slightly truncated version (15x7 = 105
days), as it is illustrated in Figure \ref{fig:Daily-arrivals-weekly}.
This means that the data matrix $Y$ was analyzed for ``weekly trends'',
employing a full-rank MF according to Eq.\ref{eq:PPCA-definition}.
Figure \ref{fig:PPCA-components-plot} illustrates these components,
which essentially are the contents of the resulting matrix $W^{H}$. 

The percentage labels on the left of each sub-plot in Figure \ref{fig:PPCA-components-plot}
represent the relative variance described by each component; in other
words, how much of the total energy of the signal is included when
it is reconstructed by each single component. Hence, each of these
``weekly trends'' constitutes a compact set of spectral descriptors
of the original signal. As in the case of SVD, using only three of
these components is enough to describe almost $\nicefrac{3}{4}$ (78\%)
of the energy content of the data series.

\subsection{ICA}

The PCA and Probabilistic PCA methods that were presented earlier
are based on orthogonal linear transformations that are error-optimal
in the MSE sense \cite{Bishop_ppca_1999}. However, there are cases
were making the transformed data statistically \emph{uncorrelated},
as the KLT does, is not adequate to create an effective mapping to
a new, more efficient space. Instead, alternative methods have to
be employed in order to make the data statistically \emph{independent}
- a much stronger requirement.

The \emph{Independent Component Analysis} (ICA) is a family of algorithms
and statistical methods for transforming a set of data into a mixture
of statistically independent components, usually in the context of
\emph{blind source separation} (BSS) tasks \cite{langers_blind_2009,bobin_blind_2008,abolghasemi_blind_2012,silva_blind_2011}.
More specifically, ICA can be viewed under the scope of MF as \cite{hyvarbook,Hyvarinen2000ica,Theodoridis.Konstantinos2009}:
\begin{equation}
Y=A\cdot S\label{eq:ICA-definition}
\end{equation}
where $Y$ is a $l\times n$ matrix of observations, i.e., $l$ measurements
for each of $n$ variables, $A$ is a $l\times k$ ``mixture'' matrix
and $S$ is a $k\times n$ matrix of $k$ statistically independent
``sources''. 

ICA has been used successfully in a wide range of data-intensive processing
tasks, from big data and data mining to fMRI unmixing \cite{bai_robust_2006,behroozi_fmri_2011,calhoun_ica_2003,calhoun_unmixing_2006}.
It is based on identifying non-Gaussian properties between the sources
and separating them from the mixture, essentially reconstructing the
original signal as a linear combination of identified components.
Naturally, the number of independent sources is bounded by the maximum
column-rank of matrix $Y$, i.e., $k\leq n$. In other words, the
original data are assumed to be generated by $k$ independent processes
from which only one may be Gaussian. The constraint of statistical
independency is defined via a non-linearity metric, usually kurtosis
or hyperbolic tangent functions.

In this study, ICA was used in a similar way as with SVD and PPCA
described earlier, i.e, the daily arrivals data series was restructured
into a weekly-grouped version of a slightly truncated version (15x7
= 105 days), as it is illustrated in Figure \ref{fig:Daily-arrivals-weekly}.
The data matrix $Y$ was analyzed via ICA for ``weekly trends'',
here in the context of generating ``sources'', i.e., weekly patterns
or ``templates'' that are statistically independent.

The ICA processing was conducted with the fastICA toolbox 2.5 for
Matlab \cite{fastica-toolbox-url}, implementing the Hyvarinen's fixed-point
algorithm \cite{Hyvarinen97afast,Hyvarinen99afast}. The fastICA has
been used in various studies \cite{correa_comparison} as a benchmark
for the ICA family of algorithms for BSS, with different choices regarding
the exact non-linearity and decorrelation approach. In this study,
all four non-linearity choices were considered, namely pow3, Gaussian,
skewness and hyperbolic tangent, since previous studies have used
different choices as optimal. Additionally, both decorrelation approaches
were used, namely \emph{symmetric} (estimate all the independent components
simultaneously) and \emph{iterative} (estimate independent components
one-by-one like in projection pursuit). In all cases, a standard PCA
pre-processing stage was included. 

Similarly to the SVD and PPCA analysis for MF, here the daily arrivals
data series was restructured into a weekly-grouped version of a slightly
truncated version (15x7 = 105 days), as it is illustrated in Figure
\ref{fig:Daily-arrivals-weekly}. The purpose here is to decompose
the data series into ``weekly trends'' via ICA, i.e., as mixtures
of statistically independent (not just uncorrelated) components, according
to Eq.\ref{eq:ICA-definition}. The significance of each individual
ICA component was investigated by means of reconstruction error, as
well as the signal energy captured by the variance in each case. 

\begin{figure}[htbp]
\begin{centering}
\textsf{\includegraphics[width=8.5cm]{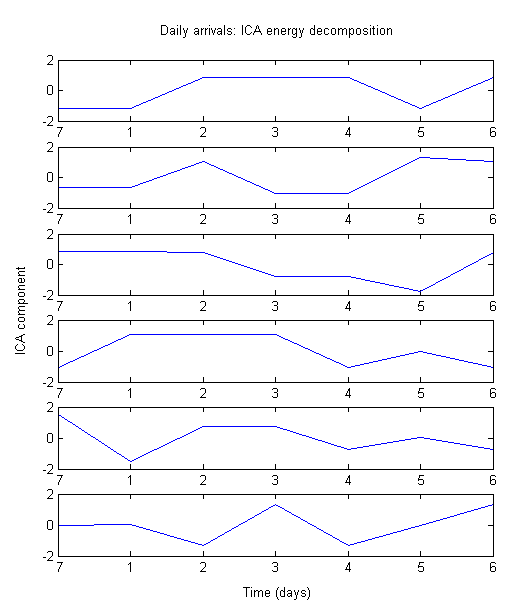}}
\par\end{centering}

\caption{\label{fig:fastICA-components-plot}fastICA components (\emph{tanh}
non-linearity, \emph{symmetric} decorrelation) of the weekly-grouped
daily arrivals; horizontal axis is the weekdays.}
\end{figure}

Since the structure of the data matrix the $Y$ in Eq.\ref{eq:ICA-definition}
hints a maximum column-rank of seven, Figure \ref{fig:fastICA-components-plot}
illustrates the first six ICA components as weekly trends, while the
seventh component is simply the residual reconstruction error. From
this plot, it is evident that there are indeed several distinct ``templates''
of weekly trends for the daily arrivals. The relevance of each one
of these patterns are quantified by the individual rank-1 reconstruction
and comparison of the resulting signal to the original data series
of daily arrivals in terms of energy. 

\begin{figure}[htbp]
\begin{centering}
\textsf{\includegraphics[width=8.5cm]{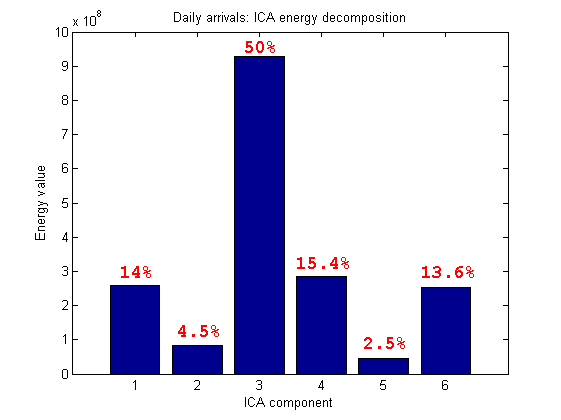}}
\par\end{centering}

\caption{\label{fig:fastICA-comp-distr-plot}Distribution of the fastICA components
(see: Figure \ref{fig:fastICA-components-plot}) of the weekly-grouped
daily arrivals, plotted against their relative (\%) spectral energy.}
\end{figure}

Figure \ref{fig:fastICA-comp-distr-plot} illustrates the distribution
of the ICA components (see: Figure \ref{fig:fastICA-components-plot})
of the weekly-grouped daily arrivals, plotted against their relative
(\%) spectral energy. It is evident that the third component (from
top) dominates the energy distribution histogram and along with the
fourth component correspond almost to $\nicefrac{2}{3}$ of the signal
energy of the original data series. Although in general the ICA components
are not directly interpretable with regard to the original domain
of the signal, these findings explain and further support the statement
that there are clear periodic trends in the daily arrivals that correspond
to short ``bursts'' and somewhat longer ``pauses'', as previous
spectral and MF methods also suggest. 

\begin{figure}[htbp]
\begin{centering}
\textsf{\includegraphics[width=8.5cm]{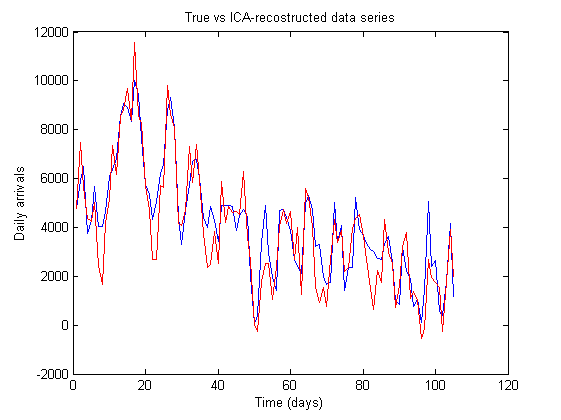}}
\par\end{centering}

\caption{\label{fig:fastICA-approx-plot}Plot of the fastICA reconstruction
of the weekly-grouped daily arrivals (red) against the true data series
(blue); the horizontal axis is the time (days).}
\end{figure}

Figure \ref{fig:fastICA-approx-plot} illustrates the true and the
rank-6 (incomplete) ICA reconstruction of the weekly-grouped daily
arrivals (red), plotted against the true data series (blue); the horizontal
axis is the time (days).

\subsection{K-SVD}

The SVD, PPCA and ICA methods for MF that were presented earlier can
all be referred to as full-rank algorithms: unless the original matrix
$Y$ is inherently rank-deficient by structure, these methods produce
a MF formulation that exploits this full-rank property, i.e., utilizes
the maximum number of components for the mixture. Hence, in the case
of the daily arrivals data series restructured here into a weekly-grouped
matrix version as in Figure \ref{fig:Daily-arrivals-weekly}, any
similar full-rank MF will produce a maximum of seven components.

A recent and very different approach to the MF problem is the introduction
of additional constraints to the task, specifically in the structure
of the components matrix. Instead of putting statistical decorrelation
or independency, the original data are formulated as a \emph{sparse}
mixture of \emph{dictionary} elements, in a similar way:

\begin{equation}
Y^{T}=D\cdot C\label{eq:DL-definition}
\end{equation}
where $Y$ is a $l\times n$ matrix of observations, i.e., $l$ measurements
for each of $n$ variables, $D$ is a $l\times k$ ``dictionary''
of $k$ elements (columns) and $C$ is the corresponding $k\times n$
matrix coefficients that define the mixture \cite{tosic_dictionary_2011,lee_efficient_2006}. 

Although the MF formulation here looks the same as in the previous
methods, Eq.\ref{eq:DL-definition} defines the mixture for $Y$ as
a product of a dictionary $D$ of components and a coefficient matrix
$C$. The idea is to use as little components from $D$ as possible
to reconstruct the original data. In practice this means that the
additional constraint is to have as many zeros as possible in every
column of matrix $C$, as this is equivalent to canceling out the
corresponding dictionary elements. For example, if $C_{*,j}=\left[0,1,0,0,-2,4,0\right]^{T}$
then only the three non-zero coefficients will be used in the mixture
(product) with the dictionary $D$ to produce the current element
(of column/variable $j$) of the data matrix $Y$. 

The dictionary $D$ may be \emph{over-complete}, which means that
at most $m\leq k$ elements may be used in the mixture but there is
no strict rank-related limit to the actual number of elements (columns)
in the dictionary, other than being able to produce a more ``packed''
representation of the original signal. This is established in practice
by combing sparsity in the coefficients and \emph{incoherency} (e.g.
decorrelation) in the dictionary elements. Although this seems equivalent
to what SVD or (P)PCA does in practice, here the incoherency metric
can be any function than compares the ``similarity'' between the
elements (columns) in dictionary $D$. Both sparsity and incoherency
constraints define the exact size $k$ in Eq.\ref{eq:DL-definition}
and they are a key property that has been studied extensively over
the last few years.

In Eq.\ref{eq:DL-definition}, the dictionary $D$ can be pre-defined
as a set of general-purpose functions that can produce an effective
spectral decomposition of the original signal - usually an over-complete
set of trigonometric of wavelet functions. This is not much different
than the standard \emph{Discrete Wavelet Transformation} (DWT), only
now there is no inherent multi-scale property and the dictionary elements
do not have to be rescaled variants of the same detail or approximation
wavelet. If properly configured, DWT may also produce a sparse or
\emph{compressed} spectral representation of a signal, however here
the sparsity constraint is explicitly defined in the decomposition
algorithm, normally via the $l_{0}$ norm or (in practice) via the
$l_{1}$ norm and variants (e.g. see: LASSO \cite{theokopsslavbookchapter}).
For proper sparse decomposition, the dictionary $D$ must also be
defined in a data-driven way, i.e, \emph{learned} by the data, instead
of being pre-defined a priori. Hence, these approaches are referred
to as \emph{Dictionary Learning} (DL) algorithms and they are currently
in the state-of-the-art in the context of the general BSS task, as
well as in Coding Theory, with a wide range of applications including
\emph{Compressed Sensing} (CS), wireless sensor networks, fMRI \&
EEG analysis, etc \cite{abolghasemi_blind_2012,abolghasemi_fast_2013,Eusipco2014,bai_robust_2012,jafari_speech_2009}.

The K-SVD algorithm is one of the most popular approaches in the standard
DL task. In practice, it implements the formulation of Eq.\ref{eq:DL-definition}
by alternating training steps of the dictionary $D$ and the coefficients
$C$. More specifically, it produces a rank-1 approximation of the
residual reconstruction error, uses it to update the matrix $D$,
then uses this new updated dictionary to produce a new mixture $C$,
repeating this cycle iteratively until a specific average sparsity
constraint is satisfied and/or the total reconstruction error becomes
smaller than a specific threshold. Due to its flexibility in terms
of defining sparsity and incoherency, K-SVD has been widely adapted
to various problem-specific tasks, e.g. allowing some limited number
of non-sparse elements in $D$ to be used in order to capture noise/artifacts,
background elements, etc \cite{Eusipco2014,rubinstein_efficient_KSVD_2008,aharon_k-svd:_2006,Lee_ddfmri_2010,rubinstein_k-svd_2012}.
Other sparsity-promoting approaches include Non-Negative Least Squares
(NNLS) approximations, without the $l_{1}$ norm \cite{skretting_recursive_2010,Bouts_NNLS_2007,Lin_NNLS_2007},
as well as special versions of ICA or DL with additional ``structural''
constraints on the produced MF approximation \cite{abolghasemi_fast_2013,adah_iva_2012,bai_robust_2006}.

\begin{figure}[htbp]
\begin{centering}
\textsf{\includegraphics[width=8.5cm]{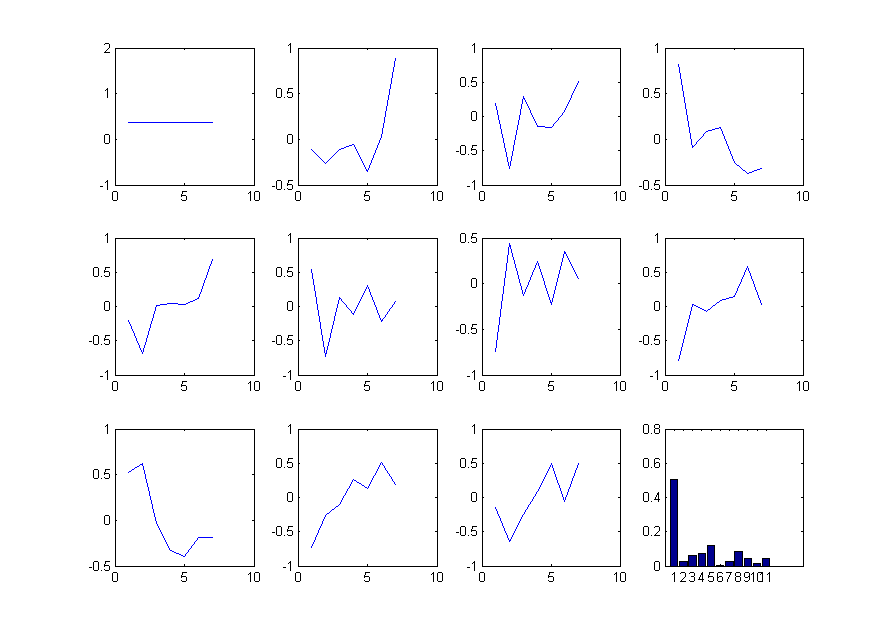}}
\par\end{centering}

\caption{\label{fig:KSVD-components-plot}K-SVD components (dictionary) of
the weekly-grouped daily arrivals; horizontal axis is the weekdays;
the histogram at the bottom-right corner is the same components ranked
according to their relative spectral energy (dictionary size = 11,
sparsity constraint = 7, reconstruction error = 2.e-12).}
\end{figure}

In this study, K-SVD was used to produce a MF formulation in the context
of DL, similarly to the ones produced by the other MF methods presented
earlier. Again, the daily arrivals data series was restructured into
a weekly-grouped version of a slightly truncated version (15x7 = 105
days), as it is illustrated in Figure \ref{fig:Daily-arrivals-weekly}.
The DL approach was employed to analyze the data matrix $Y$ for ``weekly
trends'', which are encoded as the elements of the dictionary $D$.
For practical reasons, Eq.\ref{eq:DL-definition} shows that the data
matrix $Y$ was used transposed, in order for dictionary $D$ to produce
weekly-based components. Figure \ref{fig:KSVD-components-plot} shows
the K-SVD components (columns of $D$) of the weekly-grouped daily
arrivals; horizontal axis is the weekdays; the histogram at the bottom-right
corner is the same components ranked according to their relative spectral
energy.

\begin{figure}[htbp]
\begin{centering}
\textsf{\includegraphics[width=8.5cm]{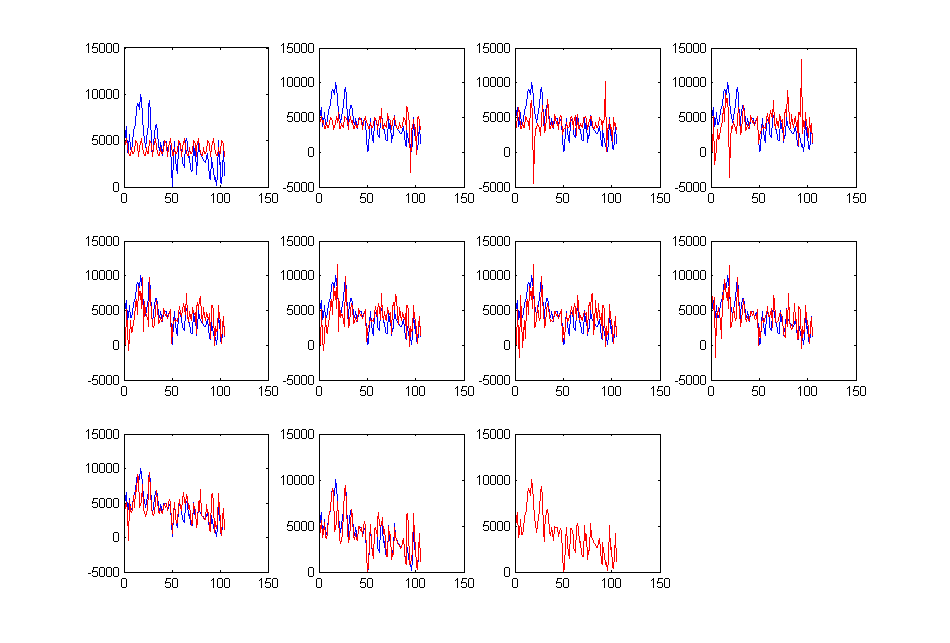}}
\par\end{centering}

\caption{\label{fig:KSVD-approx-plot}K-SVD spectral approximation (red) of
the weekly-grouped arrivals, plotted against the true data series
(blue); the horizontal axis is the time (days) (dictionary size =
11, sparsity constraint = 7, reconstruction error = 2.e-12).}
\end{figure}

Figure \ref{fig:KSVD-approx-plot} illustrates the K-SVD spectral
approximation (red) of the weekly-grouped arrivals, plotted against
the true data series (blue); the horizontal axis is the time (days).
Each individual plot corresponds to data series reconstruction that
uses one additional mixture element, i.e., from $k=1$ to $k=11$.
In each case, $m\leq\max\left\{ k,7\right\} $ corresponds to the
number of elements used from the dictionary $D$ of size $k$, which
are sorted against their overall contribution (energy) in the reconstructed
signal; in other words, each reconstruction uses a larger set of energy-sorted
elements from $D$ and, hence, produces a smaller reconstruction error.
For the maximum sparsity constraint of $m=7$ and the full dictionary
size $k=11$, perfect reconstruction is achieved.

It is clear from Figure \ref{fig:KSVD-approx-plot} that K-SVD can
produce a very accurate representation of the daily arrivals data
series by using a rank-7 mixture and at least 11 dictionary elements.
Although this seems a more complex model compared to the previous
MF approaches, these plots show that even from a dictionary of five
the reconstruction quality is already comparable to the other methods.
The reason that the reconstruction becomes perfect only when the size
becomes maximum $\left(k=11\right)$ is that the mixture is based
on a sparse representation and not all dictionary elements are used
equally. This is more evident, in a quantitative way, in the right-bottom
histogram subplot in Figure \ref{fig:KSVD-components-plot}, where
the first or ``background'' component contributes roughly 50\% of
the energy of the original signal - exactly as it was described earlier
for ICA in Figure \ref{fig:fastICA-comp-distr-plot} (third component).
However, the top-left subplot in Figure \ref{fig:KSVD-components-plot}
reveals that in this case the corresponding component is ``flat''
(constant). The remaining energy is spread in the rest of the components,
but in this case their total number is 10 instead of six and, hence,
the maximum relative contribution is roughly 10\% instead of 13-15\%
as in ICA.

\section{Fractal dimension analysis}

In recent years, dimensionality analysis in signal processing has
been extensively linked to fractal analysis and \emph{fractal dimension},
as a non-parametric method for the quantitative characterization of
the complexity or ``randomness'' of a signal \cite{fracbookg_1992,fracbookw_1994}.
When applied to 1-D signals, metrics like the \emph{Hurst exponent}
and the \emph{Lyapunov exponent} have been used as statistical features
to describe various types of data series, from biomedical signals
(e.g. EEG, ECG, etc) to financial and climate time series. In 2-D
signals, these methods provide additional features for characterizing
the texture of images, e.g. when analyzing biomedical modalities (radiology,
ultrasound, MRI, etc) \cite{chen_fracanal_1989}. Fractal dimension
is closely linked to these fractal parameters and it provides a clear
distinction between the \emph{embedding} space, i.e., the full-rank
space in the algebraic sense, from the actual space spanned by the
registered sensory data. In the general case when fractal analysis
is applied to some multi-dimensional signal, the estimation of the
fractal dimension can be used as a realistic evaluation of the ``complexity''
of the space spanned by the actual data points available and, hence,
a very useful hint regarding the inherent redundancy in a given data
set. 

In order to establish a preliminary estimation of the complexity and
intrinsic dimensionality of data sets, fractal analysis provides a
data-centric approach for this task. Data set fractal analysis, specifically
the calculation of \emph{intrinsic fractal dimension} (FD) of a data
set, provides the quantitative means of investigating the non-linearity
and the correlation between the available \emph{features} (i.e., dimensions)
in terms of dimensionality of the embedding space \cite{chen_fracanal_1989,moore_crossval_1994}. 

In the case of data series, as the daily arrivals in this study, there
are algorithms designed specifically for fast approximation of the
FD in terms of the Hurst or Lyapunov exponents. However, the most
generic approach is to treat the measurements of the data series as
an arbitrary data set with dimension of two or one, if represented
as $\left(x,y\right)$ pairs or single-valued, respectively, and process
it via dataset-oriented algorithms for estimating the FD. The two
most commonly used methods of calculating the FD in such cases are
the \emph{pair-count} ($PC$) and the \emph{box-counting} ($BC$)
algorithms \cite{fracbookw_1994,moore_crossval_1994,Pierre_fracanalimg_1996,fracbookg_1992}.
In the PC algorithm, all Euclidean distances between the samples of
the data set are calculated and a closure measure is then used to
cluster the resulting distances space into groups, according to various
ranges $r$, i.e., the maximum allowable distance within samples of
the same group. The $PC$ value is calculated for various sizes of
$r$ and it has been proved that $PC\left(r\right)$ can be approximated
by:

\begin{equation}
PC\left(r\right)=K\cdot r^{D}\label{eq:PC}
\end{equation}
where $K$ is a constant and $D$ is called PC exponent. The $PC\left(r\right)$
plot is then a plot of $log\left(PC\left(r\right)\right)$ versus
$log\left(\nicefrac{1}{r}\right)$, i.e., $D$ is the slope of the
linear part of the $PC\left(r\right)$ plot over a specific range
of distances $r$. The exponent $D$ is called \emph{correlation fractal
dimension} of the data set, or $D_{2}$. 

The BC approach calculates the exponent $D$ in a slightly different
way, in order to accommodate case of large data sets; however, it
essentially calculates an approximation of that same correlation fractal
dimension value, i.e., $D_{2}$. It is commonly used when the data
sets contain large number of samples, usually in the order of thousands
\cite{Traina_fracfeatsel_2000,Abrahao_fracdset_2003}. In this case,
instead of calculating all distances between the samples, the input
space is partitioned into a grid of $n$-dimensional cells of side
equal to $r$. Then, the samples inside each cell are calculated and
the frequency of occurrence $R_{r}$, i.e., the count of samples in
a cell, divided by the total number of samples, is used to approximate
the correlation fractal dimension by:
\begin{equation}
D_{2}=\frac{\partial log\underset{i}{\sum}\left(R_{r}^{i}\right)^{2}}{\partial log\left(\nicefrac{1}{r}\right)}\label{eq:corr-fracdim}
\end{equation}

Ideally, both PC and BC algorithms calculate approximately the same
value, i.e., the correlation fractal dimension $D_{2}$ of the initial
data set, which characterizes the intrinsic (true) dimension of the
input space \cite{Abrahao_fracdset_2003}. In other words, $D_{2}$
would be the \emph{minimum dimension of the data set} in order to
correctly represent the original data set in \emph{any} embedding
space.

In this study, FD analysis was applied to the daily arrivals data
series in the $\left(x,y\right)$ representation form, i.e., treating
the individual signal values as distribution in the full 2-D embedding
space. The PC algorithm employing Euclidean distances was used, due
to the relatively small number of samples available, as well as the
better stability and accuracy for $D_{2}$ against the BC approach
\cite{Pierre_fracanalimg_1996}. 

In order to calculate the slope at the linear part of the $PC\left(r\right)$
plot, a parametric sigmoid function was used for fitting between the
sample points of the plot. In the parametric sigmoid function: 

\begin{equation}
y=y_{0}+C_{y}\left(\frac{1}{1+exp\left(-C_{x}\left(x-x_{0}\right)\right)}\right)\label{eq:param-sigmoid}
\end{equation}
the $(x_{0},y_{0})$ identifies the transposition of the axes, while
$C_{x}$ and $C_{y}$ identify the appropriate scaling factors. Specifically,
the value of $C_{x}$ affects the steepness of the central part of
the curve, while $C_{y}$ specifies the $Y$-axis width of the sigmoid
curve. Then, the slope of the linear part around the central curvature
point, i.e. the value of $D_{2}$, is:

\begin{equation}
\frac{\partial^{2}y\left(x_{0}\right)}{\partial x^{2}}=0\Rightarrow D_{2}=\frac{\partial y\left(x_{0}\right)}{\partial x}=\frac{C_{x}\cdot C_{y}}{4}\label{eq:param-sigmoid-slope}
\end{equation}

The fitness of the parametric sigmoid over a range of samples assumes
uniform error weighting over the entire range of data. Thus, if a
large percentage of points lies near the upper bound ($y=y_{max}$)
or lower bound ($y=y_{min}$) of the $Y$-axis range, as in most cases
of $PC(r)$ plots, then the fitness in the central region of the sigmoid,
i.e., where the slope is calculated, can be fairly poor. For this
reason, an additional weighting factor was introduced in the fitness
calculation in this study. Specifically, the Tukey (tapered cosine)
parametric window function \cite{Harris_tukey_1978} was applied over
the $Y$-axis range when calculating the overall fitness error of
the sigmoid. The Tukey window is parametric ($q$-value) in terms
of the exact form around its center, ranging from completely rectangular
($q=0$) to completely triangular or Hanning window ($q=1$). When
applied over the $Y$-axis range, the rectangular case is equivalent
to calculating the fitness error uniformly over the entire range,
while the triangular case is equivalent to calculating the fitness
error primarily against the central point of the sigmoid curve. In
this study, all fitness calculations employed Tukey windows as error
weighting factors, using parameters $q$ in the range between 0.5
and 1.0 for optimal slope results. The equation for computing the
coefficients $w_{j}$ of a discrete Tukey window of length $N$ ($j=1...N$)
is as follows:
\begin{equation}
w_{j}=\begin{cases}
\begin{array}{ccc}
\frac{1}{2}\left(1+cos\left(\frac{2\pi\left(j-1\right)}{q\left(N-1\right)}-\pi\right)\right)\\
1\\
\frac{1}{2}\left(1+cos\left(\frac{2\pi}{q}-\frac{2\pi\left(j-1\right)}{q\left(N-1\right)}-\pi\right)\right)
\end{array}\end{cases}\label{eq:tukey}
\end{equation}
where the first branch is for $1\leq j<\frac{q}{2}\left(N-1\right)$,
the second for $\frac{q}{2}\left(N-1\right)\leq j\leq N-\frac{q}{2}\left(N-1\right)$
and the third for $N-\frac{q}{2}\left(N-1\right)<j\leq N$; $N$ is
the size (span) of the window, $q$ is the smoothness factor.

\begin{figure}[htbp]
\begin{centering}
\textsf{\includegraphics[width=8.5cm]{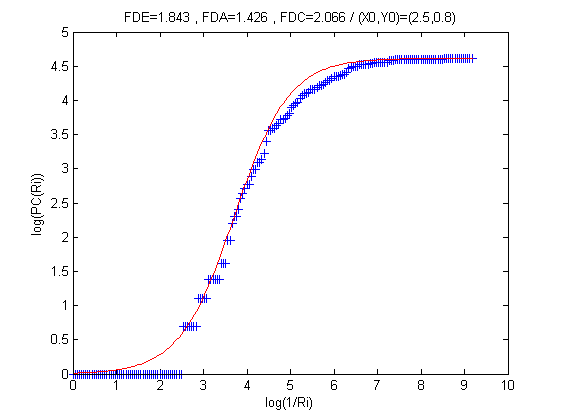}}
\par\end{centering}

\caption{\label{fig:Fractal-dim-plot}Estimation of the daily arrivals data
series FD via parametric sigmoid approximation (red) in the log-log
pair-count (blue) plot. The Tukey-weighted $\left(q=0.8\right)$ sigmoid
slope at the central point yields FD around 1.84 (``FDE'').}
\end{figure}

Figure \ref{fig:Fractal-dim-plot} illustrates the log-log PC plot
(blue) and the corresponding best-fit approximation via a parametric
sigmoid function (red). There are three different estimations that
can be used as over- and under-estimations (bounds) for the most accurate
one, namely FDA=1.43 for the complete (unweighted) sigmoid, FDC=2.07
for the central point-only and FDE=1.84 for the Tukey-weighted $\left(q=0.8\right)$
sigmoid slope at the central point. In this case, FD is expected to
be 2.0 at most (upper bound) since the original signal is structured
as 2-D; on the other hand, the further away from the value of 1, the
more stochastic (random/complex/''chaotic'') it is. Hence, the value
of 1.84 for FD is a clear hint for strong non-deterministic properties
of the daily arrivals, as expected.

\section{Discussion}

As it was mentioned from the start, the goal of this study is two-fold:
(a) the statistical and signal-level characterization of the smuggling
networks as a generating process; and (b) the draft formulation and
preliminary assessment of such models for predictive purposes, i.e.,
to produce short-term forecasting of the refugee influx. Hence, all
the results can be explained within this context of system-identification
and predictive-analytics capabilities, especially under the scope
of changes in policies and mandates that follows the ``stable''
time frame up to mid-January 2016. Predictive analytics are discussed
first, as they provide hints to the statistical properties of the
smuggling networks, followed by the system identification of the smuggling
networks, as well as some hints for future enhancements and the current
state-of-play in the refugee crisis.

\subsection{Predictive analytics of the refugee influx }

The standard statistics of the refugee influx reveal the nature of
the daily arrivals as raw data. More specifically, the first few moments
in Table \ref{tab:Stats-Basic}, as well as the histogram itself in
Figure \ref{fig:Stats-Histogram-plot}, reveal a distinct asymmetry
between the left and the right side of the Gaussian approximation.
The deviation of mean and median values from the absolute middle of
the range confirm the conclusions drawn from the values of kurtosis
and skewness, illustrating a trend towards the lower end. In practice,
influx rates lower than the mean are a bit more common and more closely
packed together than the ones higher than the mean, which are a bit
less common and more sparsely distributed (greater variation). These
observations are also confirmed by the best-fit parameters of the
corresponding Poisson and GEV models.

Based on this statistical profile, it is safe to assert that the distribution
is closely approximated by a Gaussian model, which results more or
less that it follows the 2/3 inclusion rule for the mean$\pm$stdev
range. When weekly-grouped (see: Figure \ref{fig:Daily-arrivals-weekly}),
the average daily arrivals reveal a distinct difference in volume
between the weekdays, with preference to Sundays and Mondays for the
higher influx rates - more than double than the lower influx rates,
as Figure \ref{fig:Weekday-averages} shows. Although not very useful
for actual short-term forecasting, these data-backed conclusions are
a very important verification of qualitative observations conducted
throughout the same period by rescue \& relief teams in the ``hot''
zones.

In this study, predictive analytics are described under the scope
of various approaches and algorithms, including both daily and weekly-grouped
formulations of the raw data. First, the linear and cosine-linear
regressions reveal the general trends: a downward linear trend of
47 less arrivals per day within the time frame under investigation;
and a primary periodic trend of roughly 6.2-6.5 days, i.e., a repetitive
behavioral pattern. The later is further established by examining
the largest coefficients of the (simple) linear regressor, where the
previous 1-4 \& 10-11 days seem to be the most important subset for
predicting the next day's influx rate. This periodic behavior is presented
more clearly by the spectral analysis (FFT) in Figure \ref{fig:FFT-logplot}
and the frequency-rescaling calculation, where the low-band density
profile seems bounded, again, close to the limit of six days or slightly
less than a full week. 

An actual predictive model with significant short-term forecasting
capabilities is the one described by ARMA. Different kernel sizes
and adjustments to the time reference used (here, the weekday index)
provide very useful insights of how this can be achieved with minimum
computational requirements. More specifically, an auto-regressive
convolutional kernel using no more than the previous 21 influx spot
values can produce a very close approximation to the real data series.
Most importantly, analysis of the AR coefficients prove the periodicity
of the daily arrivals and the weak correlation between days with a
lag 3-4 between them. This essentially confirms the strong correlation
between influx rates \emph{more than} three days apart, or immediately
before the current one (lag 1) as in all low-frequency signals, exactly
as the previous methods propose too.

The inherent complexity of the daily arrivals data series is characterized
quantitatively by the fractal dimension analysis that was described
earlier. In practice, a completely deterministic system would present
a linear behavior and, therefore, its embedding dimension would be
equal to one. On the other hand, a completely stochastic system would
present perfectly random fluctuations that would cover its entire
plane, i.e., its topological and its embedding dimension would coincide
to the value of two (pairs of $\left\{ x,y\right\} $ data, where
$x$ is the time index and $y$ is the value). In this study, the
fractal analysis shows that the (estimated) embedding dimension of
the influx data series is 1.84, i.e., somewhere between chaotic and
stochastic, closer to the second one. In other words, the daily arrivals
show strong non-deterministic properties, \emph{but not as much as
to make them non-predictable}, at least with regard to short-term
forecasting. 

The weekly-grouped restructuring of the daily arrivals provides an
alternative insight to the refugee influx, in terms of weekly patterns
and trends. As the results in Figure \ref{fig:SVD-approx-plot} from
SVD analysis show, the daily arrivals is a signal with strong low-frequency
signature. According to the histogram of the energy-ranked SVD components,
using just the first of them (i.e., associated with the largest eigenvalue)
is enough to capture a significant portion of the signal's energy
and, hence, its general shape. In practice, this is not much different
than employing an auto-regressive model of order seven (i.e, AR(7))
in the general sense of ARMA, in order to conduct short-term forecasting;
but in this case the model's coefficients are optimized according
to its eigenvalues via SVD. As the Figure \ref{fig:SVD-approx-plot}
shows, using the first three or four SVD components, i.e, 3x7=21 or
4x7=28 coefficients is adequate for a close approximation, similarly
to the ARMA(21,1,1) that was presented earlier (see: Figure \ref{fig:ARMA-3approx-plots}).
Hence, a three-week time window seems adequate for constructing such
analytical models for forecasting purposes in this context.

The K-SVD approach can be used in a similar way, i.e., discover weekly
patterns via SVD-based error minimization. However, this assertion
is valid only when the corresponding sparsity level is constrained
to a very small number, e.g. 1 or 2, at the cost of overall approximation
error. On the other hand, using a large dictionary and a large sparsity
level can produce a very accurate approximation of the signal, as
in Figure \ref{fig:KSVD-approx-plot}. This latter approach was the
one employed for K-SVD in this study, i.e., to illustrate how such
a MF method can be used to design an efficient predictive model for
short-term forecasting of the daily arrivals.

\subsection{System identification of the smuggling networks }

The inherent behavioral properties of the smuggling networks, operating
near the Turkish coastline and enabling the travel of thousands of
people across the sea passages to the Greek islands, are the underlying
statistics of the generating process, i.e., the ``system'' that
creates the daily influx. Normally, this could be formulated more
precisely by a full queuing model, including intermediate staging
areas or ``nodes'' inside the mainland and transition routes or
``edges'' between them, effectively creating a typical network-based
framework, e.g. for M/M/1 or M/G/1 analysis (e.g. see: \cite{Bertsekas-DataNetw-1992}).
However, this kind of in-depth analysis requires extensive data series,
not only for the daily arrivals influx to Greece, but also for every
intermediate ``buffer'' zone inside Turkey. In other words, the
current data sets are sufficient only for a ``black box'' analysis
of this generating process, namely the flow between departures (Turkey)
and arrivals (Greece).

As it was mentioned for the predictive analytics, multiple methods
point to a very clear periodic trend of roughly 6.2-6.5 days or slightly
less than a full week. Linear-cosine regression and spectral analysis
(FFT) has shown that the smuggling networks operate more or less in
identifiable, almost-weekly behavioral patterns. Additionally, basic
statistics and weekly averages show that there is a very distinct
difference in daily influx rates between the 48-hour window of Sunday/Monday
and the weekdays that follow.These are all clear evidence of a generating
process that functions, in total, as a \emph{store-and-forward} ``black
box''. It is known that the smuggling networks operating inside Turkey
can actually be viewed as \emph{flow graphs}, very similar to the
data networks that employ buffers and store-and-forward techniques
as part of their routing protocols (e.g. see: \cite{Bertsekas-DataNetw-1992}).
As described earlier, lack of detailed data for the internals of these
smuggling networks effectively means that they can only be studied
from the ``outside'', i.e., with regard to their total ``throughput''.
Nevertheless, this evidence almost certainly proves that they operate
in a (almost) two-day ``burst'' / five-day ``pause'' pattern,
an assertion that complies completely with the results from the ARMA
modeling. 

The PPCA and ICA approaches are alternatives to the MF analysis of
the weekly-grouped daily arrivals, not based on eigenvectors as in
SVD but employing decorrelation and independency as the statistical
constraints for the components, accordingly. In both cases, the blind
``discovery'' (in the BSS sense) of the underlying components or
statistical ``sources'' that characterize the generating process
is indeed a very effective investigation on how the smuggling networks.
More specifically, the dominating weekly patterns, illustrated in
Figure \ref{fig:PPCA-components-plot} for PPCA and in Figure \ref{fig:fastICA-components-plot}
for ICA, as well as their relative energy distribution (for ICA, see:
Figure \ref{fig:fastICA-comp-distr-plot}), show that not only the
signal is a low-frequency data series but the high influx rates are
mostly associated to the Sunday/Monday 48-hour time frame.

\subsection{Recent changes and retrospective }

After a long period of discussions and meetings that had started as
early as mid-November 2015, on March 18-20th 2016 the EU summit finalized
and concluded the deal with Turkey regarding the handling of refugee
flows from its coasts to the Greek islands. The mutual agreement included:
(a) Turkey's commitment in stopping the smuggling networks to drastically
reduce the refugee influx towards Greece, (b) the involvement of NATO
naval forces (SNMG2) for intensifying the monitoring of the most commonly
used sea passages and (c) the immediate registering and deportation
back to Turkey of any refugees landing in Greece, with the intent
of either granting them passage to Europe via air travel (if recognized
as ``in danger'') or sending them back to their countries of origin.

\begin{figure*}[tbph]
\begin{centering}
\textsf{\includegraphics[width=17cm]{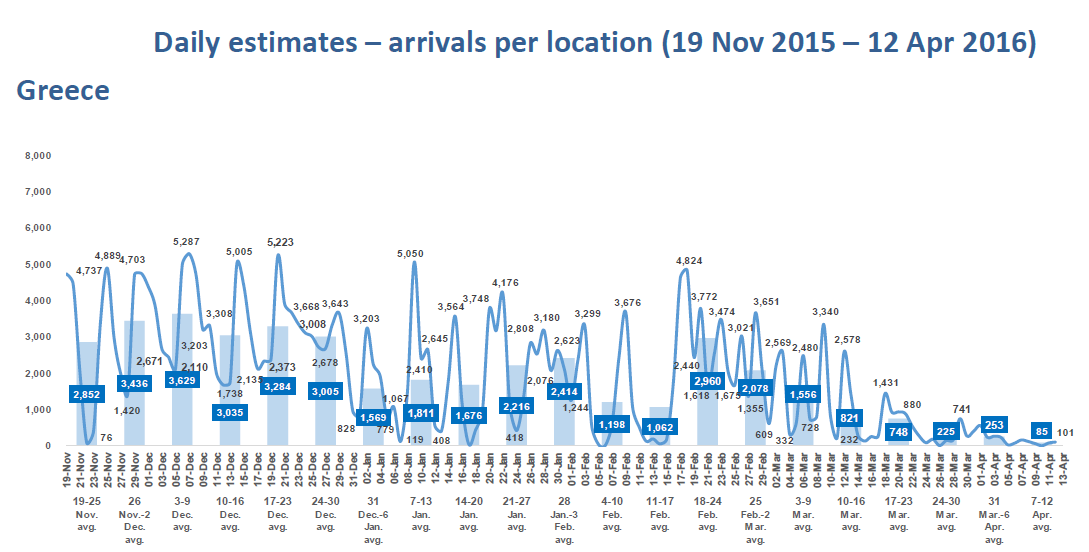}}
\par\end{centering}

\caption{\label{fig:Daily-arrivals-Greece-recent}Estimated daily arrivals
and weekly averages for the entire Greece, including the weeks just
before and after the EU-Turkey deal (Feb-Apr.2016).}
\end{figure*}

This new EU-Turkey deal had very significant and almost immediate
effects to the refugee influx in the Greek islands of first reception,
where hundreds of thousands of people had arrived during the previous
months. Figure \ref{fig:Daily-arrivals-Greece-recent} is an extension
of the data series used throughout this study, as presented in Figure
\ref{fig:Daily-arrivals-Greece}, including two special periods in
2016: (a) mid-February to March 20th and (b) March 20th to mid-April
and current status. It is clear that the plot of the first one presents
a pattern very similar to the previous period, i.e., the one analyzed
thoroughly in this study, only now the time scale seems ``compressed'':
Indeed, there are nine peaks in the daily arrivals within a month,
when the same pattern would have taken about 50 days if projected
to an earlier time - a shrinking factor of about 50/30 or 1.67 (rough
estimation, based on the plots). This can be easily explained by the
urgency of the smuggling networks to ``push'' any remaining refugee
``packets'', as fast as possible, before the EU-Turkey deal was
concluded and much stricter restrictions would be enforced in the
sea passages. With regard to the period after the March 20th, when
the deal was in place and active, it should be noted that UNHCR and
other NGOs that were involved in the reception and registration of
the refugees in Greece's ``hotspots'' decided to terminate their
presence there, stating that this deal is violating fundamental human
rights and the UN Convention about refugees and asylum seekers. As
a result, registration of any daily arrivals (significantly reduced
but certainly non-zero) is being conducted exclusively by the Greek
authorities and, hence, there are no relevant data publicly accessible,
as there was the case up to this date. Practically, this means that
(a) is a very different data series in terms of statistical properties
compared to the data set used in this study, while (b) is considered
as missing/incomplete data time frame. These limitations were the
main reason for setting the maximum usable ``valid'' date for this
study at mid-January 2016, even though more data were available. 

\begin{figure}[htbp]
\begin{centering}
\textsf{\includegraphics[width=8.5cm]{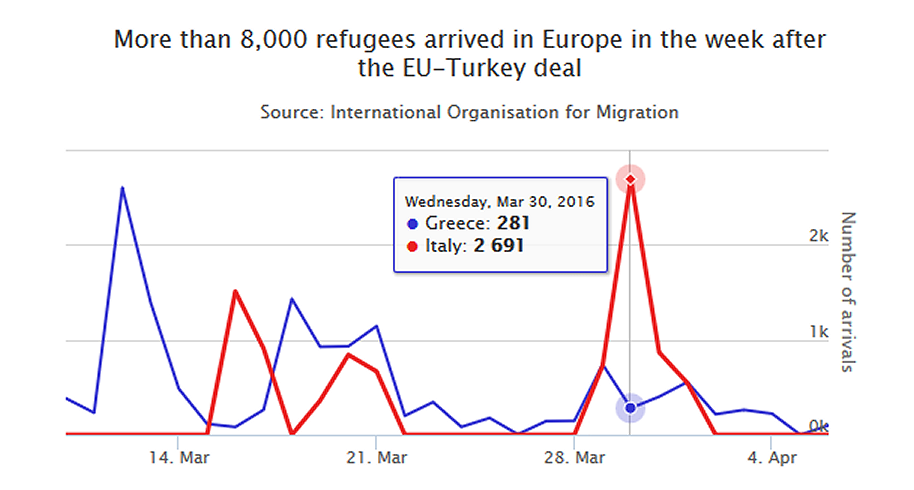}}
\par\end{centering}

\caption{\label{fig:arrivals-Greece-Italy-recent}Comparison of refugee influx
in Greece and Italy just before and after the EU-Turkey deal (Mar-Apr.2016).}
\end{figure}

Another, more important outcome from the recent EU-Turkey deal and
the strict controls imposed on the sea passages towards the Greek
islands is the gradual shift of the refugee influx to the Central
Med. route. Figure \ref{fig:arrivals-Greece-Italy-recent} illustrates
the comparison between daily arrivals to Greece and to Italy during
the weeks just before and after the deal. Although not quantified
and analyzed here, it is evident that there is a strong negative correlation
between the two data series: up to March 15th, the arrivals to Italy
were practically zero, while Greece was receiving the last large ``packets''
before the activation of the new regime; from there on, there is a
mixture of influx rates in both countries, in reduced rates; in the
week following the mark of March 20th, the influx rates are very limited
even for Greece; and finally, by the end of March there is a very
large peak of refugee influx to Italy, almost 10 times the one registered
towards Greece (2,691/281=9.57). 

The same pattern continues well within April, according to more recent
data from IOM and UNHCR (not presented here). It seems that the new
EU-Turkey deal does not actually stop the refugee influx towards Europe,
only shifts it to previous, more dangerous routes, as it was the case
18-24 months earlier when the Mare Nostrum operation (Italy) was in
place in the Central Med. route - only now, the Triton operation (EU/Frontex)
is much more limited in scope and the total refugee influx is multiplied
many times more. It is clear that, as long as the conditions remain
the same and the war zones in Syria, Iraq, Afghanistan and elsewhere
force people out and away from their homes, refugees will continue
to converge towards Europe via the Mediterranean Sea. Therefore, the
need for such early warning/alerting systems will continue to be an
imperative need in Greece, in Italy or elsewhere.

\subsection{Future work }

The current study focused entirely in the daily arrivals of the refugees,
i.e., on the influx data series. The goal was to conduct a data-driven
analysis and modeling based on this data set alone. However, it is
established that the intensity of the daily arrivals at the Greek
islands is strongly associated to specific external factors, such
as weather conditions, changes in refugee handling policies by the
EU, the intensity of fights in the war zones in Syria, etc. Some of
these factors can be quantified and included in such data-driven approaches,
others can not. 

The most promising external factor that may be used as ``input''
in these models is weather conditions. More specifically, it was pointed
out empirically from early on that some weather elements are of utmost
importance, such as wind intensity and wave height (not always correlated),
while others were of lesser importance, such as rainfall, temperature,
humidity or cloud coverage. Hence, wind intensity and wave height
are the two external factors that will be investigated subsequently,
in correlation to the core influx data series, in order to establish
the significance of the statistical relevance and use these as additional
inputs in the predictive analytics.

The issue of localization is also a factor that may be considered
in more detail. In this study, only the total influx was used as a
single data series; however, if more data are available or in cases
where the local influx rates are sufficiently high for statistical
significance (e.g. in Lesvos), localized data analysis and predictive
modeling may be conducted. Combined with localized weather information,
such systems would be even more useful and preemptive life-saving
tools in the field of SSAR operations.

In terms of an actual early warning/alerting module for integration
into rescue and relief operations, especially in ``hot'' zones like
in eastern Greece and southern Italy, these predictive analytics have
to be reformulated more strictly, assessed in terms of true performance
on unknown data (k-fold cross validation testing \cite{Theodoridis.Konstantinos2009})
and finalized with specific ``alert levels'' as output accordingly,
similar to the way such systems of Civil Protection agencies work
in other contexts, e.g. for tsunamis, wildfires, floods, etc. Ideally,
this module could be fed with live data from registration agencies
and other open data sources, publicly available Internet sources and
satellite feeds, in order to produce reliable real-time short-term
early warning of possible 24/48-hour influx periods of high intensity.
This is already investigated as an add-in feature in ``Prometheus'',
a virtual Emergency Operations Center (EOC) developed and deployed
already in Chios since January 2016 \cite{Fosscomm2016,Prometheus-url}.

\section{Conclusions}

This study presents the first-ever data-driven systemic analysis of
systemic analysis of the refugee influx in Greece, aiming at: (a)
the statistical and signal-level characterization of the smuggling
networks and (b) the formulation and preliminary assessment of such
models for predictive purposes, i.e., as the basis of such an early
warning/alerting protocol for the rescue and relief operations on-site.

The analysis employed a wide range of statistical, signal-based and
matrix factorization (decomposition) techniques. It was established
that the behavioral patterns of the smuggling networks closely match
(as expected) the regular ``burst'' and ``pause'' periods of store-and-forward
networks in digital communications. The most interesting aspect is
the discovery of a strong almost-weekly periodic trend of 6.2-6.5
days, as well as a strong preference of the Sunday/Monday 48-hour
time window for the highest peaks of influx rates. 

These results show that such models can be used successfully for short-term
forecasting of the influx intensity, producing an invaluable operational
asset for planners, decision-makers and first-responders. It is expected
that future extensions of these models, including weather factors
(wind intensity, wave height) will further increase their accuracy
and their value as actual early warning tools in such operations.

\appendices{}

\section*{Acknowledgment}

The author wishes to thank all the volunteers of the ``Platanos''
self-organized team at Skala Sykamneas, Lesvos, for their hospitality
and warm welcome in early March 2016. The embedding in their rescue,
first-response care and humanitarian relief efforts for a week has
been an invaluable insight to the situation and the real nature of
their excellent work, day and night, for countless months now. Almost
a quarter of a million refugees sailed from Turkey during 2015 across
the narrow 5 n.m. wide strip of sea and landed at the shores around
this small port. Despite the changes in policies and tactics regarding
the handling of the refugee influx towards Europe after the February
\& March 2016 deals between EU and Turkey, this has always been and
still remains one of the hottest ``front lines'' of this crisis. 

\bibliographystyle{IEEEtran}
\bibliography{refugee-game-analysis,refs}

\begin{IEEEbiography}{Harris Georgiou}
 received his B.Sc. degree in Informatics from University of Ioannina,
Greece, in 1997, and his M.Sc. degree in Digital Signal Processing
\& Computer Systems and Ph.D. degree in Machine Learning \& Medical
Imaging, from National \& Kapodistrian University of Athens, Greece,
in 2000 and 2009, respectively. Since 1998, he has been working with
the Signal \& Image Processing Lab (SIPL) in the Department of Informatics
\& Telecommunications (DIT) at National \& Kapodistrian University
of Athens (NKUA/UoA), Greece, in various academic and research projects.
In 2013-2015 he worked as a post-doctorate associate researcher with
SIPL in sparse models for distributed analysis of functional MRI (fMRI)
signals. He has been actively involved in several national and EU-funded
research \& development projects, focusing on new and emerging technologies
in Biomedicine and applications. He has also worked in the private
sector as a consultant in Software Engineering and Quality Assurance
(SQA, EDP/IT), as well as a faculty professor in private institutions
in various ICT-related subjects, for more than 18 years. His main
research interests include Machine Learning, Pattern Recognition,
Signal Processing, Medical Imaging, Soft Computing, Artificial Intelligence
and Game Theory. He has published more than 71 papers and articles
(46 peer-reviewed) in various academic journals \& conferences, open-access
publications and scientific magazines, as well as two books in Biomedical
Engineering \& Computer-Aided Diagnosis and several contributions
in seminal academic textbooks in Machine Learning \& Pattern Recognition.
He is a member of the IEEE and the ACM organizations, general secretary
in the A.C. board of the Hellenic Informatics Union (HIU), team coordinator
of the Greek ICT4D task group \& the ``Sahana4Greece'' (virtual
EOC for the refugee crisis) initiative and he has given several technical
presentations in various countries.\end{IEEEbiography}

\end{document}